\documentclass[article,twocolumn,floatfix,nofootinbib]{revtex4}
\usepackage{graphicx,amsfonts,amssymb,amsmath, hyperref, enumerate, wasysym}
\usepackage{tcolorbox}

\newif\ifhyper
\hypertrue
\ifhyper
\hypersetup{
   citecolor = {green},
   colorlinks = {true}, 
   urlcolor = {blue} 
}
\fi

\newcommand{\beq}{\begin{equation}}
\newcommand{\eeq}{\end{equation}}
\newcommand{\beqa}{\begin{eqnarray}}
\newcommand{\eeqa}{\end{eqnarray}}
\newcommand{\ket} [1] {\vert #1 \rangle}
\newcommand{\bra} [1] {\langle #1 \vert}
\newcommand{\braket}[2]{\langle #1 | #2 \rangle}

\def\bra#1{\langle#1\vert}
\def\ket#1{\vert#1\rangle}

\def\Longarrow{\protect\@lra}
\def\@lra{\relbar\joinrel\relbar\joinrel\relbar\joinrel
          \relbar\joinrel\rightarrow}

\begin{document}

\title{{ Language Design as Information Renormalization}}

\author{\'Angel J. Gallego }
\affiliation{Departament de Filologia Espanyola, Facultat de Filosofia i Lletres, Universitat Aut\`onoma de Barcelona, 08193 Bellaterra, Spain}

\author{Rom\'an Or\'us}
\email[]{roman.orus@dipc.org}
\affiliation{Donostia International Physics Center, Paseo Manuel de Lardizabal 4, E-20018 San Sebasti\'an, Spain}
\affiliation{Ikerbasque Foundation for Science, Maria Diaz de Haro 3, E-48013 Bilbao, Spain}
\affiliation{Multiverse Computing, Paseo de Miram\'on 170, E-20014 San Sebasti\'an, Spain.}

\begin{abstract}

{Here we consider some well-known facts in syntax from a physics perspective, allowing us to establish {analogies} between both fields with many consequences. Mainly, we observe that the operation MERGE, put forward by N. Chomsky in 1995, can be interpreted as a physical information coarse-graining. Thus, MERGE in linguistics entails information renormalization in physics, according to different time scales. We make this point mathematically formal in terms of {statistical} language models. In this setting, MERGE amounts to a  probability tensor implementing a coarse-graining, akin to a probabilistic context-free grammar. The probability vectors of meaningful sentences are given by stochastic tensor networks (TN) built from diagonal tensors and which are mostly loop-free, such as Tree Tensor Networks and Matrix Product States, thus being computationally very efficient to manipulate.  We show that this implies  the polynomially-decaying (long-range) correlations experimentally observed in language, and also provides arguments in favour of certain types of neural networks for language processing. Moreover, we show how to obtain such language models from quantum states that can be efficiently prepared on a quantum computer, and use this to find bounds on the perplexity of the probability distribution of words in a sentence. Implications of our results are discussed across several ambits.}

\end{abstract} 

\maketitle 

\section{Introduction}
\label{sec1}
Linguistics can be defined as ``the scientific study of language, and its form, meaning, and context" \cite{lin}. The field itself is a broad science, sometimes even a philosophy, embracing interdisciplinary ideas from a wide variety of contexts: syntax, mathematics, computer science, neuroscience... all in all, there is no common agreement concerning why human language is as it is, or even about its basic defining properties. From the point of view of Artificial Intelligence (AI), for instance, one is worried about developing accurate algorithms for speech and text recognition/prediction \cite{AI}. Additionally, the generative approach led by Noam Chomsky tries to understand the linguistic capacity from a biological perspective, as part of human cognition. As Chomsky \emph{et al.} observe \cite{Chomskyetal2017}, the point of departure is Descartes' observation that, among all animal species, only humans seem to have a language ability \cite{Descartes}. Work on comparative cognition has endorsed this insight: only humans appear to possess a mental grammar -- an ``I-language," where the ``I" stands for \emph{intensional, internal}, and \emph{individual} -- that allows us to create infinitely many meaningful expressions from a finite stock of discrete units \cite{Chomsky2017, ilang} Within the generative models, the Minimalist Program \cite{MP} tries to attribute the properties of human language to what Chomsky \cite{Chomsky2005} calls the ``third factor", namely ``to language-independent principles of data processing, structural architecture, and computational efficiency" \cite{Chomsky2005}. This picture is not different from the general study of organic systems, and D'Arcy Thompson's and Alan Turing's works on form and morphogenesis can be seen as an example \cite{morpho}.  In this framework, Chomsky proposed a basic operation, called MERGE, to build up linguistic structures \cite{merge}. In MERGE, two syntactic objects $X$ and $Y$ are combined to form a new syntactic unit $ K $, i.e., 
\beq
MERGE : X , Y  \longrightarrow K = \{X, Y\},  
\eeq
where the brackets mean that the information in $K$ is obtained from that in $X$ and $Y$. The operation can be applied recursively, thus having the ability to create different generations of units. 

In parallel to this, physics aims to understand how the universe behaves. Some of its subfields search for the fundamental mathematical laws of the building blocks of Nature, such as high-energy physics and quantum gravity. However, the knowledge of such fundamental rules (the so-called \emph{reductionism}) does not imply a priori the knowledge of the emergent laws for aggregates of many fundamental entities (the so-called \emph{emergentism} \cite{emerg}). Typical examples of this are condensed matter and solid-state physics, where the knowledge of the rules governing the fundamental entities at a short length scale (such as atoms and molecules described by Schr\"odinger's equation) does not imply, at least directly, the knowledge of the rules governing the collective behavior of aggregates at a large length scale (such as phase diagrams of matter). In famous words of P. Anderson, ``more is different"  \cite{anderson}. 

The key concept in the above discussion is that of \emph{emergence}: the collective properties of aggregates of systems may be, because of different reasons, very different from the ones of the individual systems themselves. The mathematical formalization of this paradigm in physics is achieved by the so-called Renormalization Group (RG), or simply \emph{renormalization} \cite{RG}. Originally developed (mainly) by K. Wilson and L. Kadanoff,  renormalization is  a strategy for dealing with problems at different physical scales. This scale is typically a length, energy or time scale, and allows for different descriptions of the problem as it changes. For instance, going from short to long length scales corresponds intuitively to ``zooming out" the system, effectively going from, e.g., a description of individual atoms (short scale) to a description of a solid with $O(10^{23})$ interacting atoms (long scale). At its root, a renormalization transformation is built in two steps: first, one keeps the most relevant information degrees of freedom to describe the system at a new scale  discarding the ones believed not to be relevant, and second, one implements a rescaling of the physical variable and operators/functions in order to maintain the original picture. Physics is full of successful applications of renormalization in different ambits, from condensed matter physics \cite{stat} to quantum field theory \cite{qft} and quantum information \cite{er} (each one with its own peculiarities), to the point that it has become one of the basic pillars in our current understanding of the laws of Nature. Physical theories that cannot be renormalized  are considered wrong or incomplete, {in the sense that they may lead to mathematical complications that are usually difficult to handle. As an example, the plain quantization of the Einstein-Hilbert actionin gravity is non-renormalizable, and this has been traditionally considered an important problem.}

Having said all this, our aim with this paper is twofold. On the one hand, we establish {analogies} between physics and linguistics in the context of the Minimalist Program (for linguistics) and emergence (for physics) which, once mathematically formalized, turn out to have important consequences in ambits as diverse as AI, theoretical linguistics, computer science, RNA / protein sequencing, quantum many-body systems, quantum computing, and beyond. {We make this analogy formal for the case of statistical language models.} On the other hand, we strengthen the relation between physics and linguistics, where language is the system to be understood using the machinery of physics, and of information theory in particular. 

Let us be more specific: here we observe that MERGE can be understood physically as a type of information coarse-graining according  to time scales. Roughly speaking, the linguistic information in sequences of words (short time scale) gets renormalized by successive MERGE operations up to meaningful sentences (long time scale) \footnote{A clarification is in order: here we understand ``renormalization" as the tool that allows the mathematical description of the information in a system at different physical scales,  accounted for by relevant degrees of freedom at every scale. Of course, the implementation of this idea in several contexts leads to different consequences. Well-known examples in physics are the existence of critical systems, critical exponents, universality classes, phase transitions, the $c$-theorem, the reshuffling of Hilbert spaces, the $\beta$-function, fixed points, RG flows, scaling laws, relevant / irrelevant / marginal perturbations... the list is unending. In our case, however, we do not assume necessarily the existence of any of these in the case of language (though some of them may also be there), and adopt instead the most fundamental and general perspective of what ``renormalization" means at its very basic core at the level of information.}. This simple observation, somehow trivial from a physics perspective, turns out to have deep consequences. In particular we show that \emph{language models} (i.e., probability distributions over word sequences) \cite{lanmod}, widely used in AI applications, admit a natural description in terms of Tensor Networks (TN) \cite{tn}. For instance, the simplest MERGE corresponds to a $3$-index tensor of components $M_{\alpha \beta \gamma}$ accounting for a probability distribution of three variables $\alpha, \beta$ and $\gamma$. And this is nothing but a \emph{Probabilistic (or Weighted) Context-Free Grammar} (PCFG), in a way to be made precise later. Probabilities of meaningful sentences with a given syntax tree are naturally given in this framework by \emph{(mostly) loop-free TNs} { made of diagonal tensors.} Such mathematical structures have a number of nice properties which make them particularly amenable to manipulations of their information content, as we shall explain. { Moreover, we show that they naturally encompass the long-range correlations that are observed experimentally in language, as well as provide arguments in favour of certain types of neural networks for language processing.} Then, we move on to show that the TN structure and the particularities of PCFGs allow for the description of the probability distributions in terms of some  \emph{TN quantum states}. Such an exotic description using quantum mechanics is only to be understood at the practical level, but it happens to provide a useful connection between computational linguistics and quantum information and computation, opening the door to unprecedented results and developments in language processing algorithms.  As examples of this, we show how such states can be built efficiently on a quantum computer, and prove lower bounds on the \emph{perplexity} of the probability distribution of a set of words in a sentence \footnote{To be defined in Eq.(\ref{perp}).} by using mathematical tools borrowed from from the theory of quantum many-body entanglement \cite{nielsenchuang}. We envisage consequences of our results in several ambits.  For instance, one can use the full machinery of TNs and quantum information to validate, simulate, assess, and improve current language models. Moreover, the fact that such probabilistic models can be fed into a quantum computer means that we have, in fact, a quantum algorithm that allows perfect random sampling of language, which is simply impossible with classical computing. All in all, and together with other implications, we propose that our physical picture is in fact related to the conjectured ``perfect design and economy" of language in Chomsky's Minimalist Program, as well as to the (also conjectured) efficient processing of linguistic information in the human brain \cite{effbrain}. 

\begin{figure}
	\centering
	\includegraphics[width=0.5\linewidth]{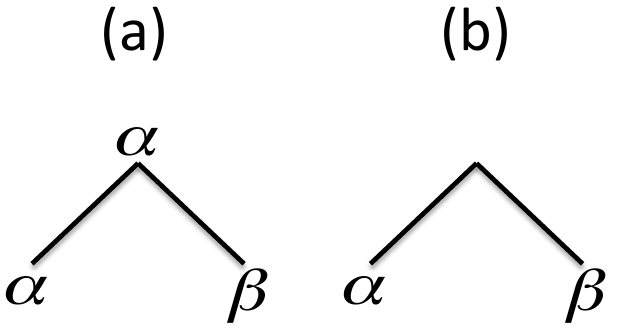}
	\caption{MERGE operation, taking two lexical elements $\alpha$ and $\beta$, and projecting them into a new one, namely $K$, with label $\alpha$. The fact that the label is also $\alpha$ means that the element resulting from the projection has the syntactic properties of $\alpha$ (the ``head" of the syntactic object). (b) A label-free representation of the application of MERGE, compatible with the recent  claim \cite{Chomsky2013} that labels should be dispensed with.}
	\label{fig1}
\end{figure}

\begin{figure}
	\centering
	\includegraphics[width=1\linewidth]{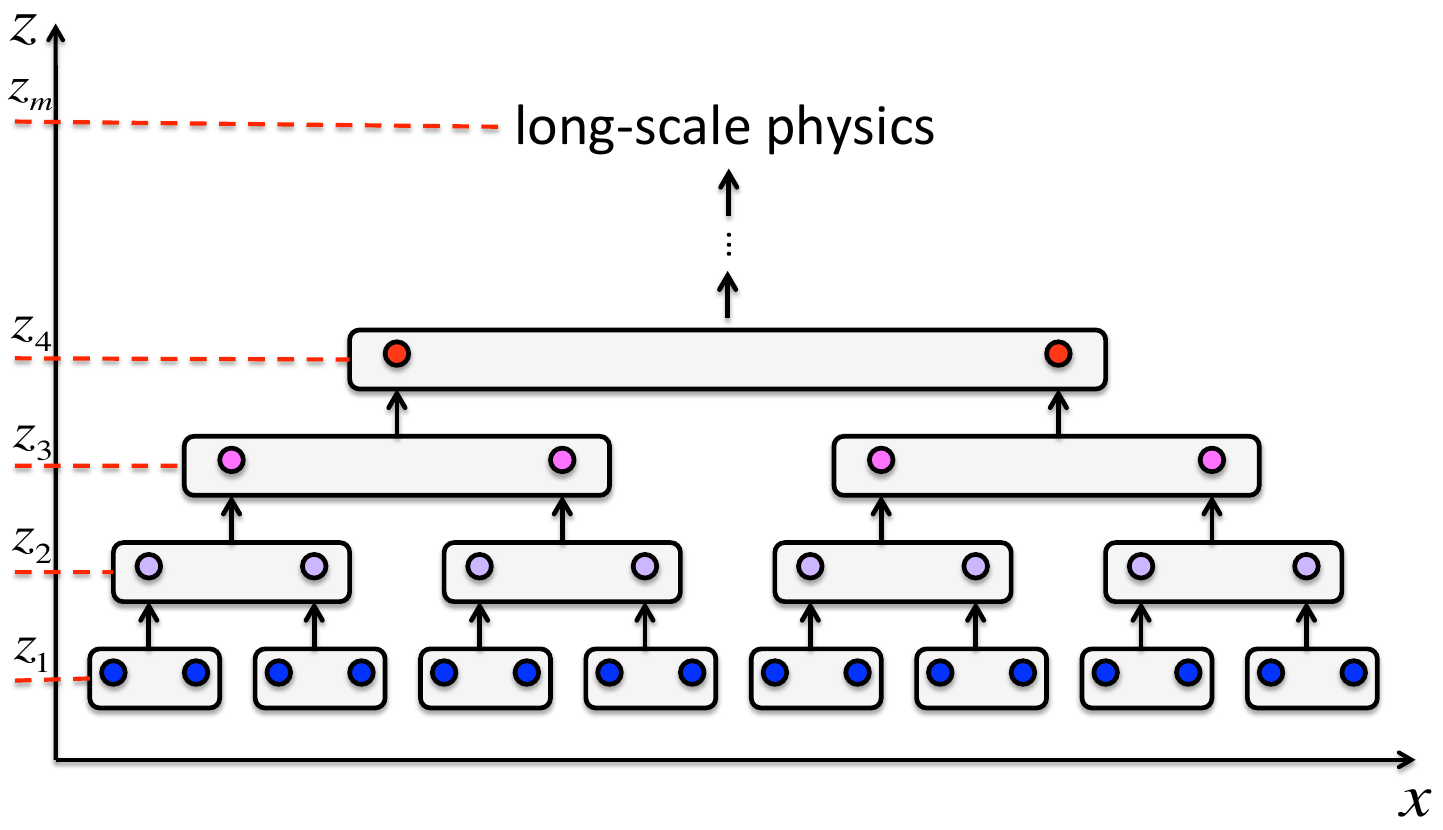}
	\caption{(Color online) Pictorial representation of a renormalization process in real space for a $1d$ lattice. The horizontal axis is coordinate $x$ (say, a space coordinate), and the vertical axis is coordinate $z$, which parametrizes the renormalization scale. Short renormalization scales (small $z$) amounts to a microscopic description of the system at small distances in $x$, whereas large renormalization scales (large $z$) amounts to a coarse-grained, macroscopic description of the relevant physics of the system at large distances in $x$. We codify short scales with ``blue" and long scales with ``red", following the intuition in physics that renormalization may take you from high energies (ultraviolet) to low energies (infrared). In our case, though, the colors have no special meaning and are just a convenient way of indicating the different scales $z_1, z_2, ...$, which are also shown for convenience. Formally, an RG step amounts to a coarse-graining followed by a rescaling of the lattice and associated operators/functions, which we implicitly assumed in the picture.}
	\label{fig2}
\end{figure}

{
The structure of this paper is as follows. In Sec.\ref{sec2} we introduce our basic {analogy}, namely, that MERGE in linguistics entails information-renormalization in physics. In Sec.\ref{sec3} we explain a direct consequence of this: the quasi-loop-free TN structure of language models. We also derive properties of such structures using tools from TNs. In Sec.\ref{newsec} we prove that our formalism encompasses the observed long-range correlations in language. In Sec.\ref{qcomp} we establish the novel connection to TN quantum states, and show in Sec.\ref{secperplexity} how this can be used to derive results on the perplexity of probabilistic language models. Then, Sec.\ref{depen} considers arbitrary grammars and language models. In Sec.\ref{sec4} we discuss some implications of our observations in different ambits.  Finally, in Sec.\ref{sec5} we wrap up our conclusions, include a table of the main {analogies} discussed in the paper, and discuss future perspectives. We also include Appendix \ref{appa} with formalities for the readers with background on theoretical linguistics, and which allows us to find even more {analogies} between linguistic and physical concepts, all of them linked to MERGE and renormalization. Overall, though, the paper is written assuming that the reader has mostly a physics $\&$ maths background, even though the style is highly heterogeneous, being this a consequence of the interdisciplinary nature of our results. }

\section{The basic {analogy}}
\label{sec2}
Our guiding principle is the following {analogy}, which we write in the form of an equation: 
\begin{center}
\centering
{MERGE  $=$ Coarse-graining}
\end{center}
The left hand side of the above expression is a purely linguistic concept. MERGE is a basic operation in syntax, picking up a set of linguistic elements (such as lexical categories) and returning a new element describing the main features of the combination of the original, see Fig.\ref{fig1}. On the other hand, the right hand side of the equation is a purely physical concept. A coarse-graining of information means the removal of superfluous degrees of freedom in order to describe a given physical system at a different scale (of, e.g., length, energy or time), see Fig.\ref{fig2}. Combined with rescaling, it is the procedure entailing renormalization, by which the rules describing the macroscopic \emph{emerge} from those describing the microscopic. The above equation establishes that both concepts are, in fact, the same basic idea but in different contexts. Chomsky's MERGE operation entails then the renormalization of linguistic information. Moreover, this renormalization accounts for different \emph{time scales}. 

\begin{figure}
	\centering
	\includegraphics[width=0.5\linewidth]{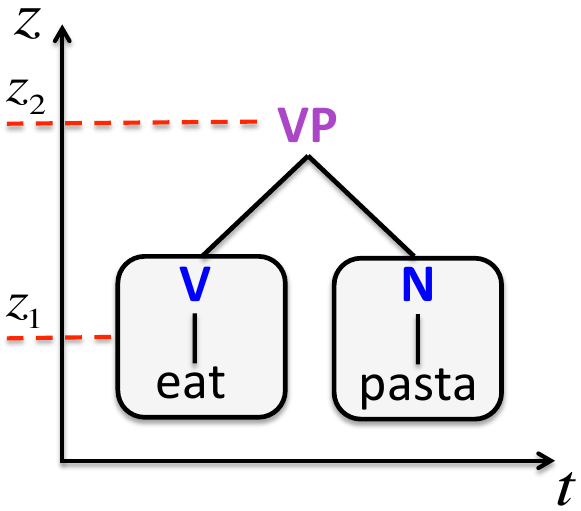}
	\caption{(Color online) The linguistic MERGE operation, seen as a physical coarse-graining process. The horizontal axis is time $t$, and the vertical axis is the renormalization scale $z$. In this case, the operation takes a verb (eat) and a noun (pasta), and coarse-grains them into a verb phrase (eat pasta). At the scale $z_2$, all the relevant syntactic information is that the compound object is a verb phrase ($VP$). Unless stated otherwise, we assume that the basic building blocks are words together with their label, as shown in the grey boxes, though one could also interpret them more fundamentally as the result of a MERGE operation between a word and a set of lexical categories. For simplicity, we shall also assume here that no other information is carried by MERGE (such as genre, number, case, etc). In any case, this extra information can always be accounted for with minor trivial modifications of the scheme that we present here. The diagram provides the structure of \emph{linguistic} correlations in the \emph{physical} $\langle z,t \rangle$ plane.}
	\label{fig3}
\end{figure}

\begin{figure}
	\centering
	\includegraphics[width=1\linewidth]{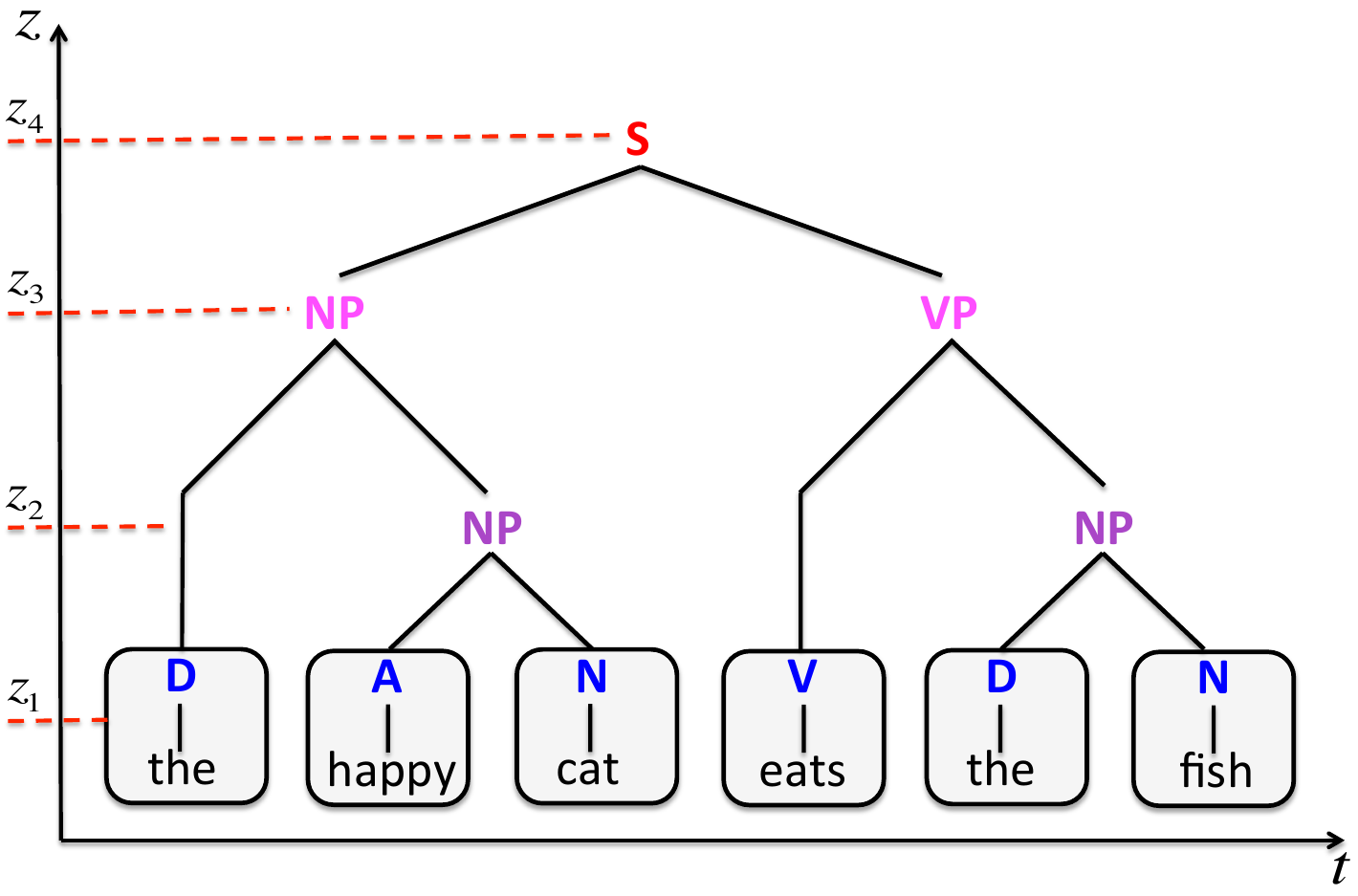}
	\caption{(Color online) Syntax tree for ``The happy cat eats the fish", seen as a renormalization process. The flow in $z$ goes from the individual words, to the sentence, labeled by $S$. The different labels correspond to the different types of syntagmas (Noun Phrase, Verb Phrase, and so on). A rescaling of the time variable at every scale is also implicitly assumed.}
	\label{fig4}
\end{figure}

To understand better why this is the case, see the example in Fig.\ref{fig3}. In terms of information processing, the MERGE operation picks up two information units at a given time scale, namely 
\beq
[_V ~ {\rm eat} ] ~~ {\rm and}   ~~ [_N ~{\rm pasta}], 
\eeq
and keeps for the next time scale the most relevant information of their \emph{coarse-grained} combination, i.e., that 
\beq
[_{VP} ~ [_V ~ {\rm eat} ] ~ [_N ~ {\rm pasta} ]]
\eeq
is a verb phrase (i.e., lexical category $VP$), where we used bracketed phrase markers to represent the syntax tree. There is information loss, since at the coarsed-grained level ($VP$) we do not have, a priori,  the information about the individual constituents (in this case, $V$ and $N$). The operation MERGE is non-associative, as corresponds in general to a coarse-graining \footnote{Meaning that $[ \alpha, [\beta, \gamma]] \neq [[\alpha, \beta], \gamma]$ using the bracketed notation.}. Moreover, as linguists know very well, grammar rules for individual words are not the same as those governing more complex syntagmas such as noun and verb phrases. So, we have two different descriptions of a system at different scales, and with different linguistic information units. The physical variable with different scales must be time, since language is spoken and thought as time-ordered sequences of concepts, being written language just a graphical representation of this, see the more complex example in Fig.\ref{fig4}. {Let us also stress that here we did not consider any semantic unit of information, but rather focus only on the syntaxis. This is the approach that we take in this paper, for the sake of simplicity of explanations. However, it would not be difficult to expand our ideas here also to include semantic information, so that sentences such as ``Colorless green ideas sleep furiously" are ruled out naturally on semantic grounds.}

This observation is ubiquitous in syntax and, when seen from the perspective of physics, entails the renormalization of linguistic information at different time scales. One can start by renormalizing words into sintagmas, then sintagmas into more sintagmas... an so on. Consider for instance syntax trees (or parse trees, as known in computational linguistics) like the one in Fig.\ref{fig4}. Such analysis are of the kind linguist use to describe how different elements (words) come together in order to produce a meaningful (active or passive) sentence, and have since long been widely used in the study of language. In practice, such syntax trees are nothing but the concatenation of several MERGE operations at different scales \footnote{Other operations can be accounted for by introducing extra links in the graphical representation, as we shall explain, but the renormalization picture still holds.}: from words to other sintagmas, from these sintagmas to more complex sintagmas... and finally up to a sentence. According to our basic {analogy}, the syntax tree  that one obtains from such analysis is nothing but the graphical representation of the renormalization of the (linguistic) information of a sentence. This is, how the information in different words comes together, hyerarchically at different time scales, up to an emergent meaningful sentence that we can interpret semantically. 

Moreover, the syntax tree also encodes the \emph{structure of physical correlations at different time scales} in the sentence. More precisely, because of the local nature of MERGE, \emph{correlations in a sentence are (essentially) built locally at different time scales}. Of course, it could be possible that other potentially-necessary operations in syntax, different from MERGE, introduce other dependencies (e.g., long-range movement). But still, it should be possible to codify them pictorically in the syntax tree. Worst-case scenario, such extra elements would introduce some loop in the  tree. But even in such a case, the renormalization picture still holds, as we shall show in explicit examples.   

Importantly, this observation is completely general, and therefore \emph{must hold for any reasonable model of language following the Minimlalist Program}. In particular, theoretical linguistic models trying to account for the observed rules of grammar, as well as probabilistic language models in artificial intelligence accounting for the chances of finding a given sentence in a corpus, should somehow encompass the renormalization of linguistic information. This, in turn, has deep implications in the structure of correlations in syntax. As we shall see in the next section, a direct consequence of our observation is a natural description of probabilistic language models in terms of quasi-loop-free TNs \cite{tn} accounting for different time scales. 
        
\section{Tensor networks and probabilistic language models}         
\label{sec3}         
        
We now consider the implications of our basic {analogy} for language models, i.e., probability distributions over sequences of words \cite{lanmod}. Such models produce probabilities $p_{w_1, \ldots, w_n}$ for a sequence of $n$ words, represented by the random variables $w_1, \ldots, w_n$, and are widely used in several technological ambits such as speech recognition, machine translation, text prediction, and so forth. In practice, such probabilities are obtained by \emph{training} the model (in this  context, computing the frequencies of sequences) with very large corpuses of text. A priori, the structure of the probability distribution depends on the grammatical constraints imposed by language, yet usually one assumes different types of ansatz. Here we focus on the general constraints that renormalization imposes on the structure of these probability distributions. As we shall see, a very natural description in terms of TNs just pops out, linking directly to Probabilistic Context-Free Grammars (PCFG), but not necessarily restricted to them only. 

\subsection{The MERGE tensor}

To begin with, let us consider the probability distribution that two given linguistic elements $\alpha$ and $\beta$ (e.g., two words) merge into a new element $\gamma$ (e.g., some other syntagma). This probability distribution $M([\alpha, \beta] \rightarrow \gamma) = M(\alpha \cap \beta \cap \gamma )$ can in fact be described by a probability map $M$, 
\beq
M : V_{in_1} \otimes V_{in_2} \longrightarrow V_{out}, 
\label{mergeeq}
\eeq
with $V_{in_1}, V_{in_2}$ and $V_{out}$ the input and output vector spaces. The coefficients of this map are given by a 3-index probability tensor $M_{\alpha \beta \gamma}$. The entries of this tensor are the probabilities of merging $\alpha$ and $\beta$ (the linguistic input of MERGE) into $\gamma$ (the linguistic output of MERGE). Physically, the tensor coarse-grains the variables $\alpha$ and $\beta$, at a given time scale, and retains the fundamental degrees of freedom of the common object at a different time scale. The result of this coarse-graining is variable $\gamma$. 

\begin{figure}
	\centering
	\includegraphics[width=0.9\linewidth]{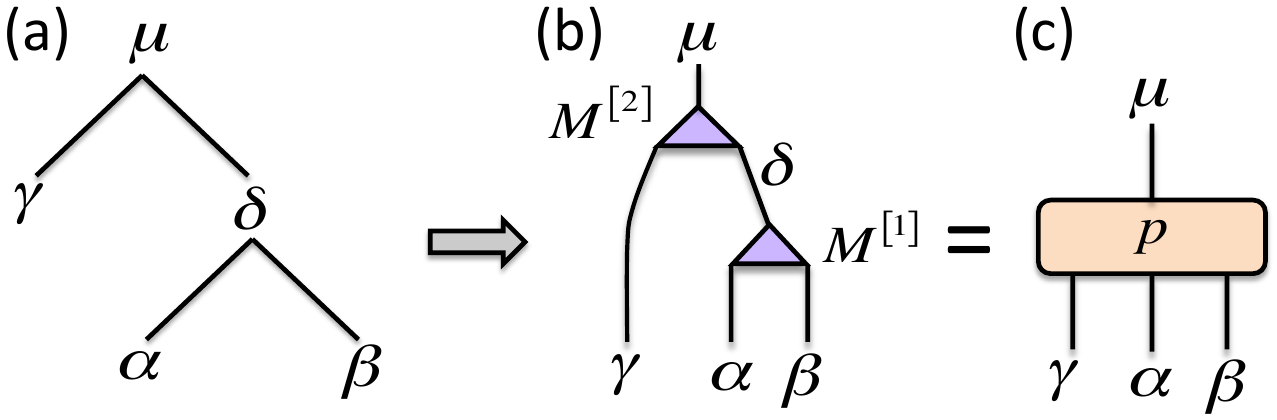}
	\caption{(Color online) (a) Two concatenated MERGE operations, where we write different greek letters for all the possible lexical variables. For language models, this structure can be represented by the tensor network in (b), where $M^{[1]}$ and $M^{[2]}$ are two different MERGE probability tensors (see text). The contraction of the tensor network gives the probability tensor $p_{\mu \gamma \alpha \beta}$, see Eq.(\ref{eqp}). In this picture we used a diagrammatic notation for tensors and their contractions, see text.}
	\label{fig5}
\end{figure}

The tensor $M_{\alpha \beta \gamma}$ obeys the usual normalization condition for probabilities, 
\beq
\sum_{\alpha, \beta, \gamma} M_{\alpha \beta \gamma} = 1, 
\eeq
i.e., the sum of all the probabilities is equal to 1. One can also compute residual probability distributions in the usual way, i.e., by summing up  over the variables that are discarded. For instance, one could have 
\beq
M'_{\gamma} = \sum_{\alpha, \beta} M_{\alpha \beta \gamma},  
\eeq
with $M'_{\gamma}$ the residual probability distribution of obtaining $\gamma$ as the output of MERGE, no matter the input. 

From a linguistic point of view, the tensor $M_{\alpha \beta \gamma}$ is the implementation, at a mathematical level, of the MERGE operation for a probabilistic language model. If the same tensor is to be used everywhere in a syntactic structure, then this is nothing but the realization of a PCFG \cite{pcfg}, i.e., a Context-Free Grammar with probabilities assigned to its merging rules. From the perspective of physics, though, this tensor coarse-grains degrees of freedom $\alpha$ and $\beta$ at a given time scale into a new degree of freedom $\gamma$ at a different time scale. From a mathematical perspective, this tensor describes the probability of obtaining the information codified in $\gamma$ from the information in $\alpha$ and $\beta$. Regardless of the interpretation, this object constitutes the fundamental LEGO\textsuperscript{\textregistered} brick of probabilistic language models following the Minimalist Program.    

\begin{figure}
	\centering
	\includegraphics[width=1\linewidth]{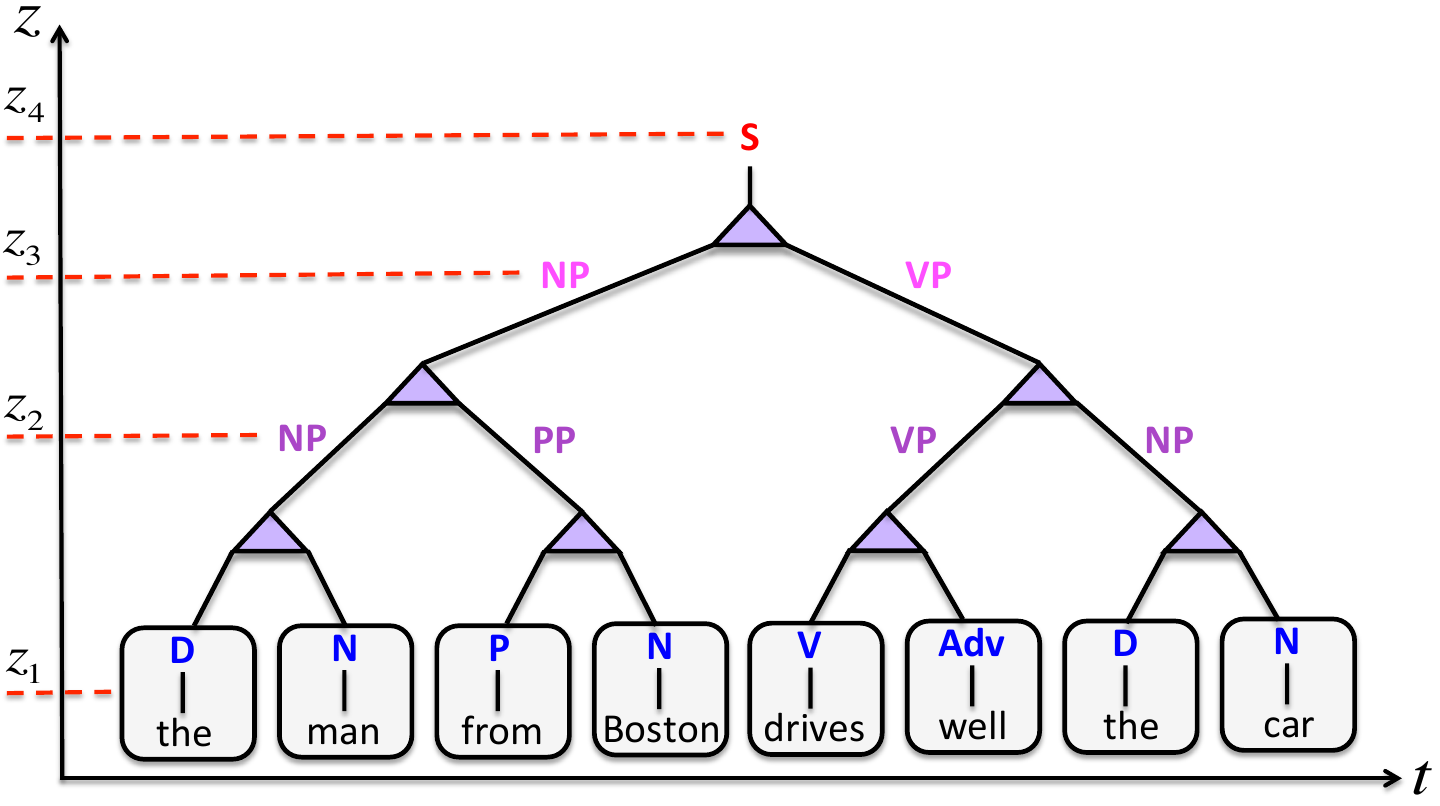}
	\caption{(Color online) Syntactic TN for the sentence ``The man from Boston drives well the car", where we included also the $t$ and $z$ axis, as well as the different renormalization scales. Linguistic information is naturally encoded in the TN at every possible scale. The contraction of the TN gives the probability of this sentence. In this particular example, the TN is a (binary) Tree Tensor Network. {Notice that the usual construction in English would be ``The man from Boston drives the car well". Our construction here is, however intentional, in order to exemplify better a perfect binary Tree Tensor Network structure.} }
	\label{fig6}
\end{figure}

\subsection{Syntactic tensor networks}
Next, we notice that the structure of a syntax tree maps directly into a \emph{tensor network} (TN) \cite{tn} for the probability distribution $p_{w_1, \ldots, w_n}$ of the sentence. Specifically, every syntactic MERGE$^{[i]}$ corresponds to a 3-index tensor $M^{[i]}_{\alpha \beta \gamma}$, with $i$ simply a label to identify individual tensors, which could in principle be different. Now let us consider the case in which a variable $\mu$ is the result of merging $\delta$ and $\gamma$, with $\delta$ itself being the result of merging $\alpha$ and $\beta$. In such a case, following the usual mathematical treatment of probabilities, one has that the probability of obtaining $\mu$ from $\alpha, \beta$ and $\gamma$ (i.e., no matter the value of $\delta$) is given by the expression
\beq
p_{\mu \gamma \alpha \beta} = \sum_\delta M^{[2]}_{\mu \delta \gamma} ~ M^{[1]}_{\delta \alpha \beta},
\label{eqp}
\eeq
i.e., we sum over all the possible intermediate events represented by $\delta$. This admits a very intuitive diagrammatic representation, see Fig.\ref{fig5}. In that figure, every tensor is a shape and every index is a line. Open indices, i.e., those over which there is no sum, are just ``free" lines, whereas sums over all the possible values of a common index between tensors are represented by lines connecting the tensors. Such sums are called \emph{contractions}, i.e., in this example we just contracted index $\delta$. These type of structures, where one has a set of tensors whose indices are contracted according to some network pattern, are called \emph{tensor networks} (TN) \cite{tn}, and always admit a convenient diagrammatic representation as in Fig.\ref{fig5}. With this in mind, we arrive to the conclusion that syntax trees of sentences map into TNs of MERGE tensors $M^{[i]}_{\alpha \beta \gamma}$ at the level of probabilistic language models. We call such structures \emph{syntactic TNs}. 

\begin{figure}
	\centering
	\includegraphics[width=1\linewidth]{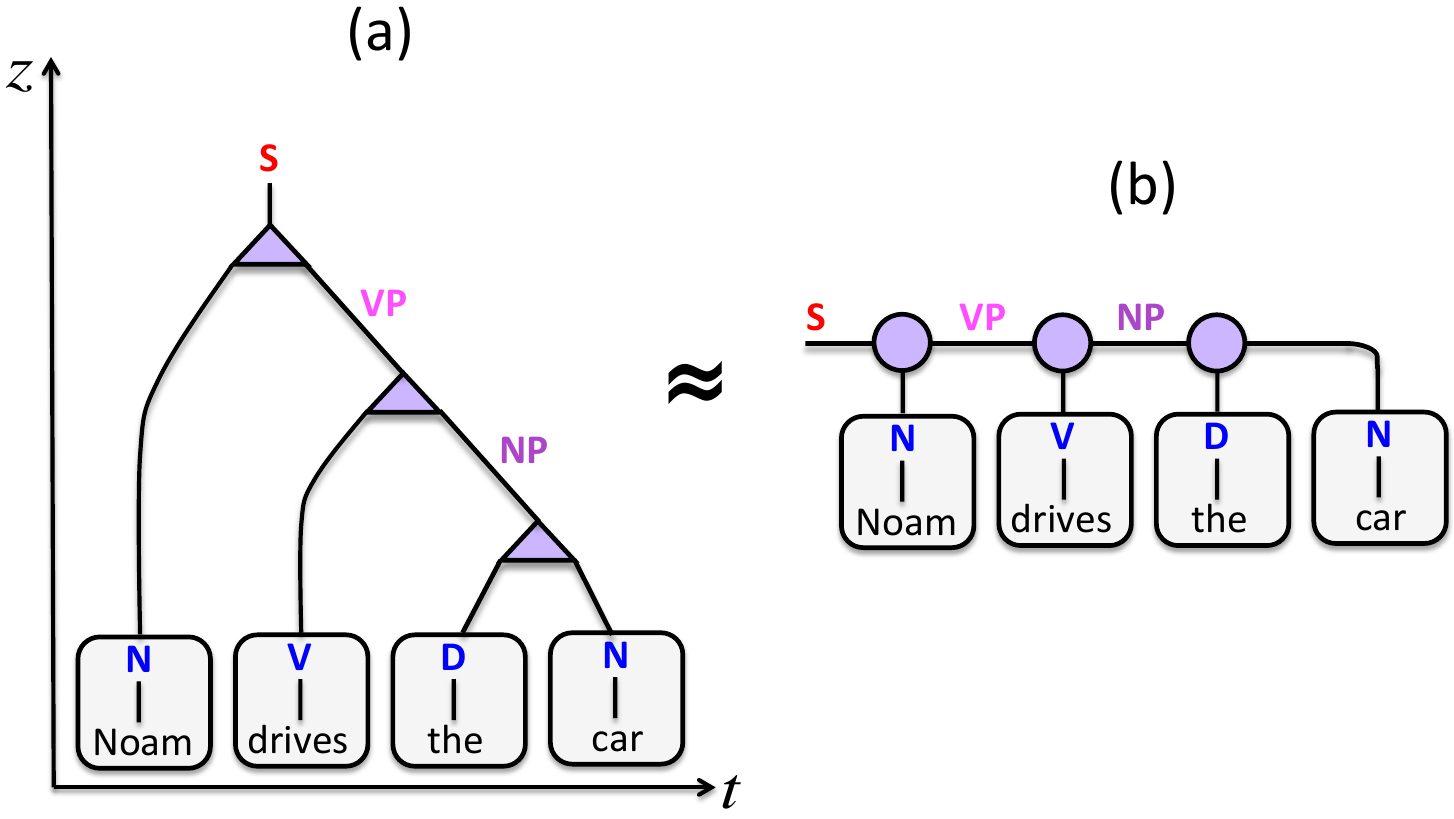}
	\caption{(Color online) Syntactic TN for the sentence ``Noam drives the car". The Tree Tensor Network in (a) can be understood as a Matrix Product State, as shown in (b).}
	\label{fig7}
\end{figure}

\begin{figure}
	\centering
	\includegraphics[width=1\linewidth]{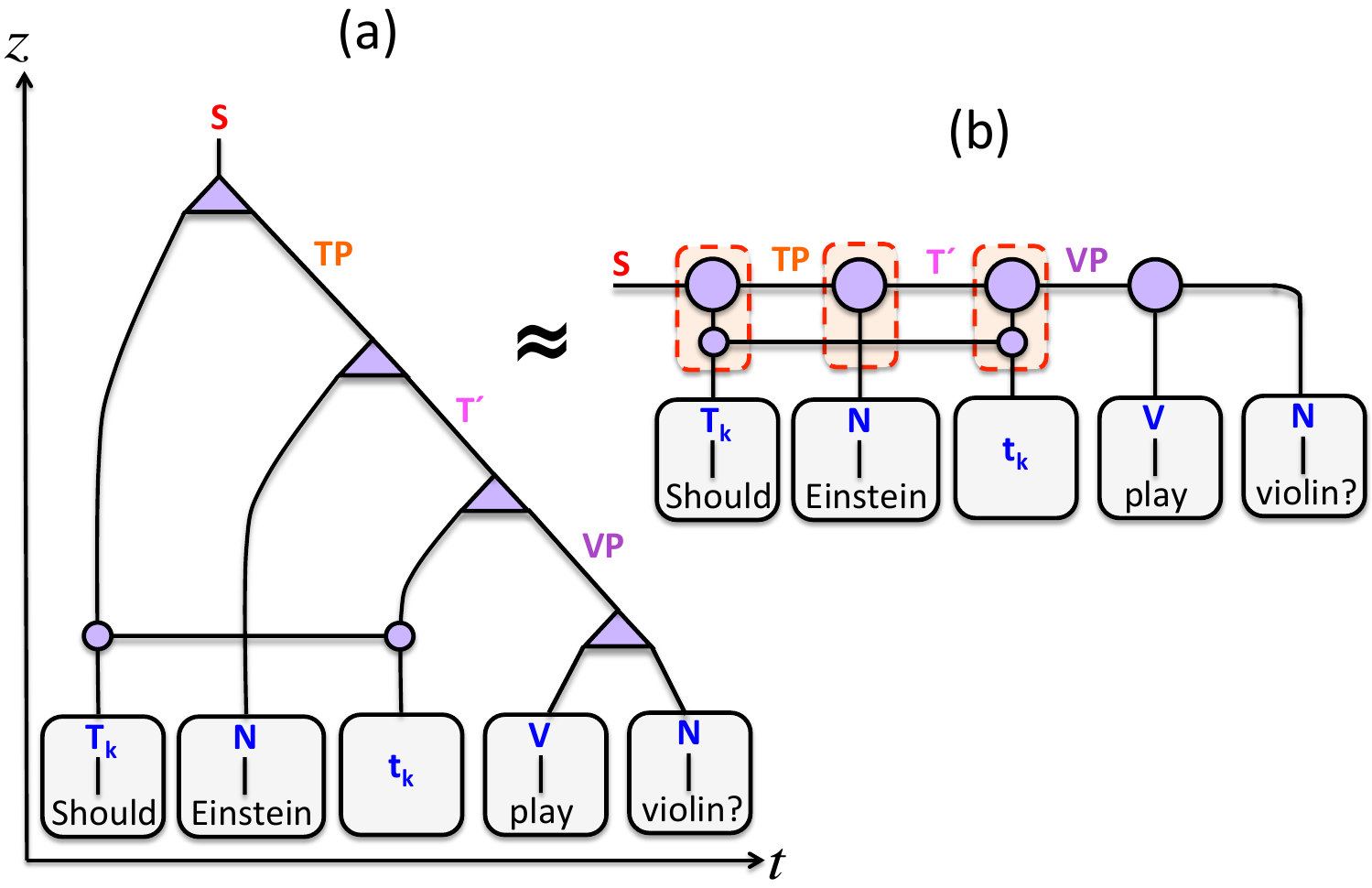}
	\caption{(Color online) Syntactic TN for the sentence ``Should Einstein play violin?", as an example of syntactic movement. The element ``Should" is created at the position of $t_k$ but externalized at the position of $T_k$ (hence it ``moved"). At the level of the TN, this can easily be accounted for by an extra correlation between these two positions, i.e., an extra link between them (and perhaps two new tensors, as shown in the figure). This introduces a loop in the TN. However, as shown in (b), it is possible to redefine the overall structure as a loop-free TN with tensors as those shown in the dotted red boxes, and reshaped (or fused) tensor indices (i.e., whenever there are two indices together, fuse them into a single big index).}
	\label{fig8}
\end{figure}

Let us be more precise: if the syntax tree does not have long-range dependencies (i.e., it is made only of MERGEs), then the TN is loop-free and corresponds generically to a Tree Tensor Network (TTN) \cite{ttn}, see Fig.\ref{fig6}. If the MERGEs are sequential in time, then the TN is in fact a special case of TTN called Matrix Product State (MPS), see Fig.\ref{fig7} \cite{mps}. These two types of structures appear quite often in the study of strongly correlated classical and quantum lattice systems in one spatial dimension \cite{tn, ttn, mps} as well as in tensor calculus \cite{tcal}, and their properties are very well known by physicists and mathematicians. Moreover, if the syntax tree has some long-range dependency (e.g., movement, agree, c-command...), then this introduces some extra index in the network, correlating variables at different positions, and therefore introducing some loop in the diagram. To be precise, such extra index \emph{correlates} the (perhaps distant) probability distributions for such variables, and can normally be casted into redefined tensors in order to keep the overall tree structure, as shown in the figure. As an example, this is in fact the case of the so-called CHAINS, which we mentioned in the introduction (Sec.\ref{sec1}), and where a lexical object is intrinsically interpreted in different contexts of a sentence but only externalized in one of them, see Fig.\ref{fig8} for an example. More intricate cases, such as those involving a concatenation of chains (the so-called \emph{successive cyclicity}), can also be accounted for similarly, see Fig.\ref{fig8b} for an example. At any rate, though, the number of loops in the TN is always quite small, as long as the syntax tree is based on a Phrase-Structure (Constituency) Grammar \cite{psg}, such as PCFGs. For the sake of clarity we restrict our explanation to these grammars. Other plausible situations, such as those arising in Dependency Grammars \cite{dependen}, will be briefly discussed in Sec.\ref{depen}.         

The syntactic structure of a sentence implies, therefore, that correlations in its probability distribution are orchestrated according to a (mostly loop-free) TN of MERGE tensors, which organize the degrees of freedom according to different time scales.  

\subsection{Properties} 

Let us now enumerate some important properties of the probability structures that we just found, which come out naturally from their TN description. Some of them were already mentioned briefly, but we revisit them again for clarity:

\subsubsection{Locally-built syntactic correlations at every scale}
\begin{figure}
	\centering
	\includegraphics[width=1\linewidth]{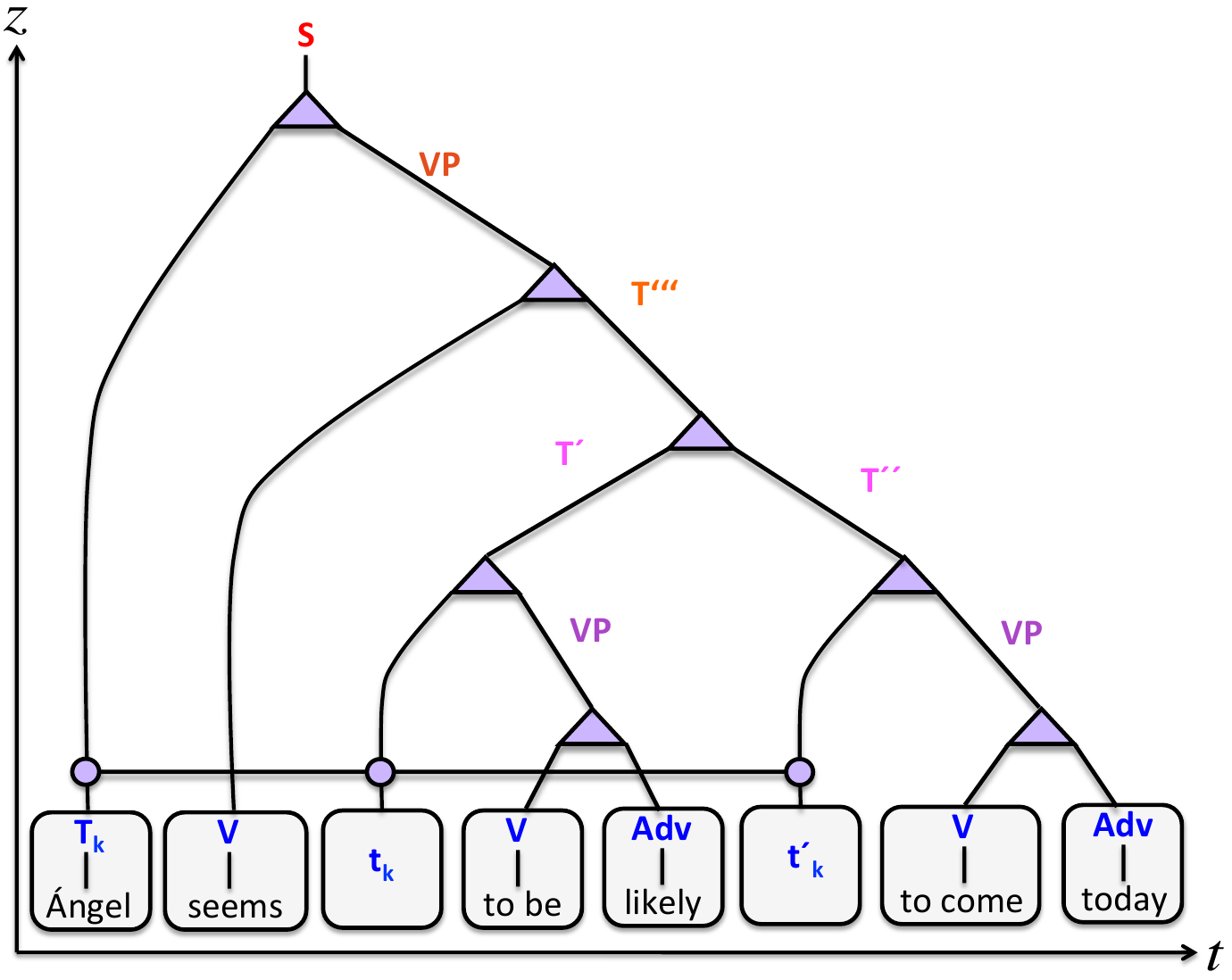}
	\caption{(Color online) Syntactic TN for the sentence ``\'Angel seems to be likely to come today", as an example of concatenation of chains, or successive cyclicity. The syntactic information of the element ``\'Angel" is at different places leaving traces $t_k, t'_k, ...$, but externalized at only one position $T_k$. At the level of the TN this can be easily accounted for by an extra index correlating different sites, similarly as in Fig.\ref{fig8}.}
	\label{fig8b}
\end{figure}

Correlations in the probability distribution are built locally at every renormalization time scale by MERGE. Distant parts of the sentence become correlated at long time scales (i.e., up in the syntax tree), whereas those that are close become correlated at short time scales (i.e., down in the syntax tree). This locality implies a nice feature of loop-free syntax trees: for a sentence with $n$ words, there are always exactly $n-1$ merged objects. Translated into syntactic TNs, this means that if the TN has $n$ indices at the first renormalization scale $z_1$ (i.e., those corresponding to the words in the sentence), then there will be exactly $n-1$ indices on the whole at higher renormalization scales $z_m, m > 1$. This can be easily checked by inspection in all the figures with loop-free syntax trees and syntactic TNs of this paper. The consequence is that to specify the full syntax of a typical sentence of $n$ words, one requires on the whole $2n-1$ units of syntactic information. In the case of having TNs with loops, as in the case of long-range movement in Fig.\ref{fig8} and Fig.\ref{fig8b}, the index creating the loop establishes a correlation between distant positions in the sentence, some of them with only syntactic information and no word present (the so called \emph{traces}). In such cases it is clear that the number of required syntactic information units is larger than $2n-1$, though not much larger.

\subsubsection{Diagonal tensors and correlated factorization}
The TN, when contracted from up towards down, reproduces the different probability distributions of the linguistic variables at every renormalization time scale. In other words, the TN encodes the probabilities of the relevant degrees of freedom \emph{at all possible time scales}. Moreover, it is possible to obtain the residual probability of \emph{any} of the variables just by contracting all the rest. Quite importantly, in syntactic TNs one does not even need to perform any tensor contraction since, once the sentence is fixed or partially fixed, there is a \emph{correlated factorization} of the whole TN because of the way human language turns out to be, which we explain in what follows. 

A well-known fact in grammar is that the output of a MERGE operation is always uniquely determined by its input. This is, given two objects being merged, there is only one possible output, no matter the context. This is a simple observation about how human language seems to work: the human brain does not merge an adjective $A$ and a noun $N$ into an object that sometimes behaves like a noun phrase $NP$, and sometimes like an adjectival phrase $AP$. Instead the combined object behaves \emph{always} like a noun phrase $NP$. So, given the input of MERGE, its output becomes fixed uniquely \footnote{Notice that the converse is not true.}. 

\begin{figure}
	\centering
	\includegraphics[width=0.5\linewidth]{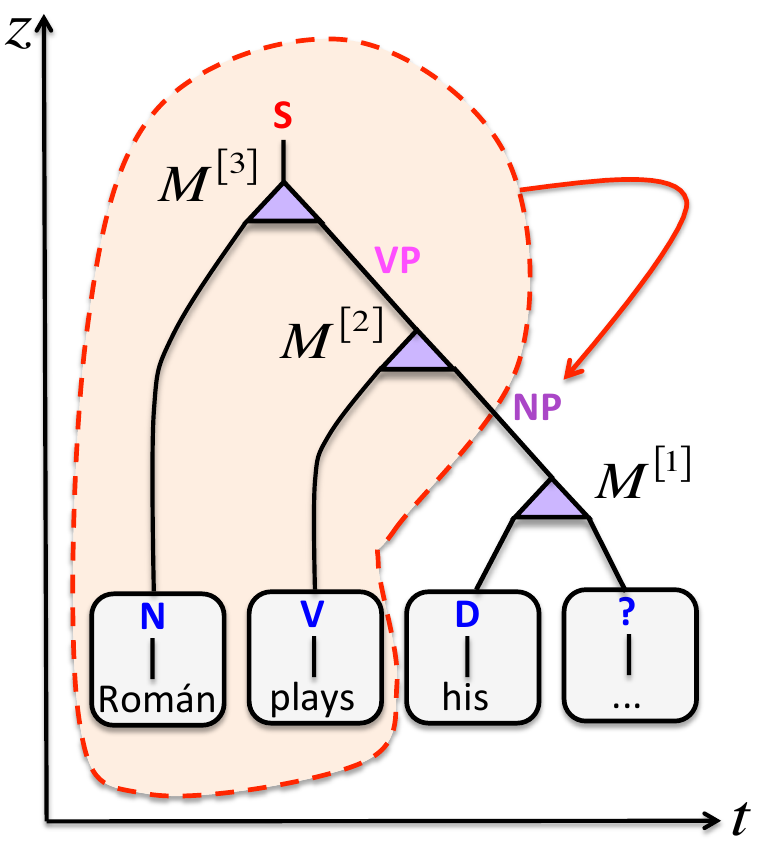}
	\caption{(Color online) Syntactic TN for the sentence ``Rom\'an plays his ...", where the last word is unspecified. The syntactic environment inside the dashed area forces the upper index of tensor $M^{[1]}$ to be $NP$. The first index of $M^{[1]}$ is forced to be the determiner ``his". This constraints the probability of finding a given word at the last place of the sentence: whatever it is, it needs to merge with a determiner to become a noun phrase. There are not too many options: the word needs to be a noun. Notice that this is fully determined by the immediate neighbourhood in the sentence (the determiner), as well as the syntactic environment (the dashed region).}
	\label{fig9a}
\end{figure}

This turns out to have an important consequence for us: { MERGE tensors are \emph{diagonal}.} As a consequence, once the sentence is given, or partially given, then the TN factorizes in a correlated way. To see why this is so, notice that if the output of MERGE is always uniquely determined by its input, then all the indices in the syntactic TN \emph{become fixed once the indices at the shortest time scale are fixed}, i.e., once a specific sentence is given. Because of this, the probability of a specific sentence actually factors out in terms of correlated probabilities and no TN contraction is needed at all. The overall correct syntactic structure of the sentence is the global, non-local property that correlates all the probabilities amongst themselves. Moreover, the residual probability of, say, finding a specific word in a sentence that is partially given, can be easily computed using one MERGE tensor only, which contains information about both the immediate neighborhood of the word, as well as about the overall syntactic neighborhood, see Fig.\ref{fig9a}. This is a very remarkable property that has its roots in the peculiarities of human language. In particular, it implies that the calculation of probabilities is \emph{extremely} efficient, and that if the correct syntactic structure of a sentence is fully or partially known, then the \emph{statistical perplexities} of reduced probability distributions are remarkably low, as we shall discuss in more detail in the forthcoming sections.  

\begin{figure}
	\centering
	\includegraphics[width=1\linewidth]{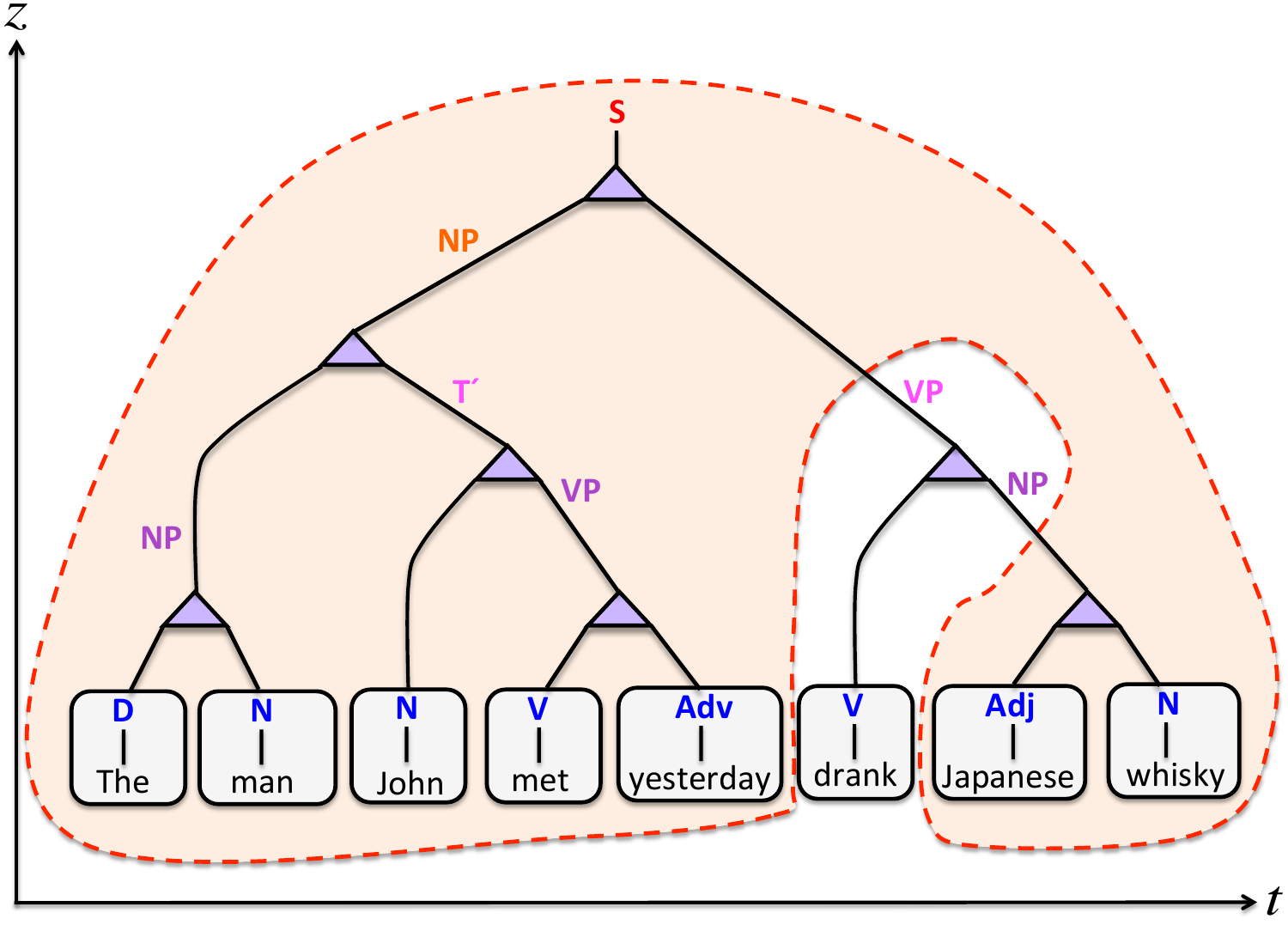}
	\caption{(Color online) Syntactic TN for the sentence ``The man John met yesterday drank Japanese whisky". The full syntactic environment of the word ``drank" is highlighted in the dashed region, and determines the probability distribution of finding a specific lexical element at that place.}
	\label{factor2}
\end{figure}

For a given sentence, therefore, the formalism produces a correlated structure of $3$-index tensors linking all possible renormalization scales, see Fig.\ref{fig9a}. For example, the overall probability of, e.g., the $4$-word sentence ``Rom\'an plays his guitar" (an actual possibility in Fig.\ref{fig9a}) reads
\beq
p_{w_1^*,w_2^*w_3^*,w_4^*} = M^{[3]}_{w_1^*, VP, S}  M^{[2]}_{w_2^*, NP, VP} M^{[1]}_{w_3^*, w_4^*, NP},   
\eeq
where $w_1^*, ..., w_4^*$ are the fixed words of the sentence, and no tensor contraction is needed at all. The above equation is a correlated product of coefficients from $3$-index probability distributions, which encode \emph{all} the syntactic information of the sentence at all time scales. The effect of this is more dramatic when it comes to residual probabilities: consider for instance predicting the word ``drank" in the sentence ``The man John met yesterday drank japanese whisky". A $3$-gram model \cite{ngram} (a rather common option in speech recognition) would give a probability distribution such as  
\beq
p_{w_4^*, w_5^*, w_6} ~~~~~ {\rm 3-gram ~ model}, 
\eeq
i.e., correlating the word $w_6$ only to ``met" and ``yesterday". The predictive power of this distribution is thus not very good, because there is no use whatsoever of the syntactic information from the rest of the sentence. However, in our TN description, the residual probability distribution, as shown in Fig.\ref{factor2}, is given by 
\beq
M^{[6]}_{w_6, NP, VP} ~~~~ {\rm Syntactic ~ TN ~ model}, 
\eeq
which includes all the relevant syntactic information of the environment needed to predict $w_6$ in the sentence. In other words, having $[_{NP} ~ [_A ~ {\rm japanese} ] ~ [_N ~ {\rm whisky} ]]$, the rest of the sentence imposes that whatever goes in $w_6$ needs to combine together with this $NP$ necessarily into a verb phrase $VP$. To put it simply, the marginal probability distribution is governed by this question: with whom do I merge to form what, as constrained by the rest of the sentence? In hindsight, this description includes all the relevant syntactic information  required to predict the word exactly at that point.  

From the above derivations, it is clear that all probabilities can be computed very efficiently, and exactly, from the TN. To be more specific, the fact that the structures are mostly loop-free implies that the calculation of probabilities, which amounts to the contraction of the tensors in the TN, can be done in polynomial time in the number of words, i.e., $O({\rm poly}(n))$ \cite{tn, ttn, mps}. From the perspective of complexity theory, this is a consequence of the fact that the contraction of a loop-free TN is a problem in the complexity class P \cite{papadimitrou} \footnote{Unlike some TNs with loops, which are $\sharp$P-hard and therefore need an exponential time to be contracted \cite{nphard}.}. But in the case of syntactic TNs like the ones described here, the situation is even better because of the correlated factorization explained above, which implies that no contraction of tensors needs to be done at all. The calculation of the probability of a given sentence amounts, simply, to determining the relevant syntax tree for a sentence and then multipliying the corresponding MERGE coefficients. For a sentence with $n$ words, it is easy to see that both steps have a computational cost of $O(n)$, and therefore the overall cost is also $O(n)$. Therefore, the renormalization structure imposed by MERGE implies a \emph{very} economical manipulation of the linguistic information in terms of computational resources such as time, memory, and energy. 

\subsubsection{Syntactic correlations}
The two-point correlations in the probability distributions depend on the sentence (specifically, the syntax tree) and the renormalization time scale chosen to compute the correlations. This is also a well-known property of  loop-free TNs, and in our case it means that the correlation between two words in the sentence decays exponentially fast with their \emph{separation distance in the syntax tree} (i.e., their separation in the network), \emph{which may be equal to the actual separation distance in the sentence or not}.

\begin{figure}
	\centering
	\includegraphics[width=0.6\linewidth]{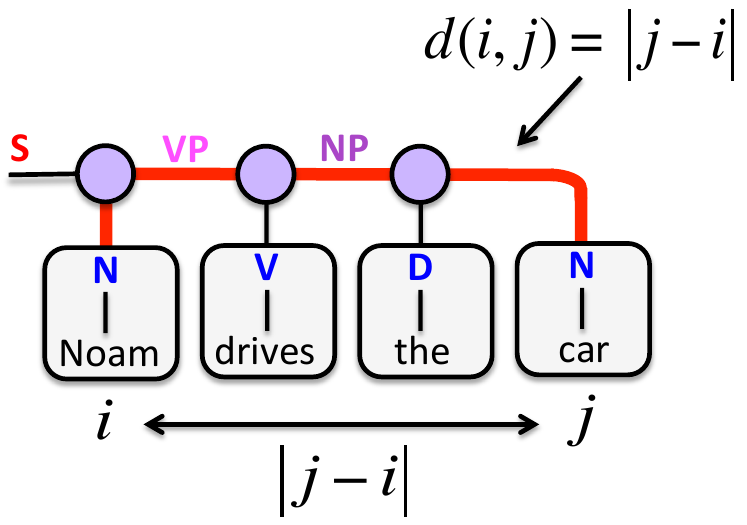}
	\caption{(Color online) For a TN structure such as the one for ``Noam drives the car",  the syntactic distance $d(i,j)$ is the same as the time separation distance, i.e., $d(i,j) = |j-i|$. This is because the structure of correlations can be written as a Matrix Product State. Two-point correlation functions in this type of sentences decay exponentially fast in the time separation $|j-i|$, as explained in the text. The syntactic path between $i$ and $j$ is shown with a red thick line.}
	\label{fig10}
\end{figure}

\begin{figure}
	\centering
	\includegraphics[width=1\linewidth]{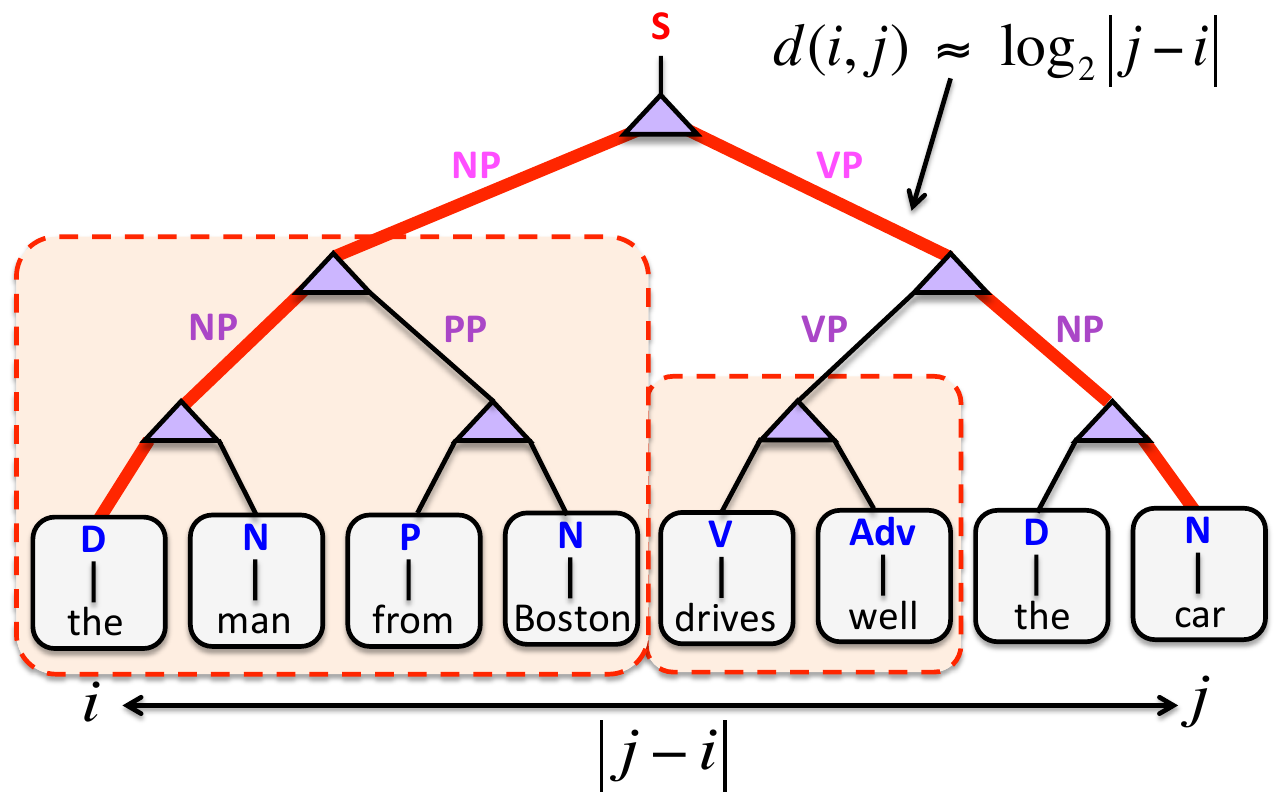}
	\caption{(Color online)  For a TN structure such as the one for ``The man from Boston drives well the car", the syntactic distance $d(i,j)$ is \emph{not} the time distance, but rather its logarithm, i.e., $d(i,j) \approx \log_2  |j-i|$. This is so because the syntactic path between positions $i$ and $j$ \emph{goes also through the renormalization scale}. Consequently, there are two-point correlation functions for these types of sentences which can decay polynomially fast towards zero in the time separation $|j-i|$, hence much slower than in the case of Fig.\ref{fig10} (see text). The path between $i$ and $j$ is shown with a red thick line. Notice, however, that at the level of the renormalized syntagmas in the red dotted boxes, the structure is exactly the same as the one in Fig.\ref{fig10}.}
	\label{fig11}
\end{figure}

Mathematically, this means the following: consider the two-point correlation function  
\beq
C(i,j) \equiv \langle f(w_i) f'(w_j) \rangle - \langle f(w_i) \rangle \langle f'(w_j) \rangle, 
\eeq
with 
\beqa 
\langle f(w_i) f'(w_j) \rangle &=& \sum_{w_1, \cdots, w_n} f(w_i) f'(w_j) ~ p_{w_1, \ldots, w_n} \nonumber \\
\langle f(w_i) \rangle &=& \sum_{w_1, \cdots, w_n} f(w_i)  ~ p_{w_1, \ldots, w_n} \nonumber \\
\langle f'(w_j) \rangle &=& \sum_{w_1, \cdots, w_n}  f'(w_j) ~ p_{w_1, \ldots, w_n}, 
\eeqa
and $f(w_i), f'(w_j)$ some functions of the variables $w_i, w_j$. We could think of these variables as those representing words at times $i$ and $j$, but they could also be the variables for other (renormalized) syntagmas at a longer time scale (i.e., somewhere up in the tree). It is possible to prove mathematically \cite{tn, ttn, mps} that this correlation function decays asymptotically as 
\beq
C(i,j) \approx e^{-d(i,j)/\tau} ~~~ {\rm for} ~~~ d(i,j) \gg \tau ,
\label{de0}
\eeq
with $d(i,j)$ the size of the path between $w_i$ and $w_j$ in the syntax tree, and $\tau$ a sentence-dependent (finite) correlation time, see Fig.\ref{fig10} and Fig.\ref{fig11}. As is well known from the theory of TNs, parameter $\tau$ does not depend on the choice of functions $f(w_i)$ and $f'(w_j)$, so it depends \emph{only} on the type of sentence and the MERGE probabilities. This conclusion also holds if the TN has a small number of loops.  Importantly, the quantity $d(i,j)$ can depend a lot on the type of syntax tree that one has. Consider for instance the two examples ``Noam drives the car", and ``The man from Boston drives well the car", with syntax trees as in Fig.\ref{fig10} and Fig.\ref{fig11}. In the first case, Fig.\ref{fig10}, the syntax tree is purely sequential, and therefore the TN for the probability distribution is a Matrix Product State \cite{mps}. In such a case it is clear that the distance $d(i,j)$ between two words is the actual separation distance in the sentence, i.e., $d(i,j) = |j-i|$. However, in the second case, Fig.\ref{fig11}, the syntax tree is a binary tree, and therefore the corresponding TN is a Tree Tensor Network \cite{ttn}. In such a case, the path along the tree between two words in the sentence \emph{necessarily goes also along the vertical axis}, and one can prove that it is given by $d(i,j) \approx \log_2 |j-i|$, again with $|j-i|$ the separation in the sentence, and where $\approx$ means that it is correct up to some possible additive constant term \footnote{To be precise, this is correct in average, since depending on the tree it is possible to choose specific pairs of points with longer separation \cite{ttn}.}. Therefore, in cases such as the one in Fig.\ref{fig10} (``Noam drives the car"), the correlation function between two words will behave like 
\beq
C(i,j) \approx e^{-|j-i|/\tau} ~~~ {\rm for} ~~~ |j-i| \gg \tau ,
\label{de1}
\eeq
whereas in cases such as Fig.\ref{fig11} (``The man  from Boston drives well the car") it will behave like 
\beq
C(i,j) \approx e^{-(\log_2 |j-i|)/\tau} \approx \frac{1}{|j-i|^{1/\tau}} ~~~ {\rm for} ~~~ |j-i| \gg \tau .
\label{de2}
\eeq
In both cases the correlation falls down towards zero with the separation distance $|j-i|$ in the sentence, but in the first case it decreases exponetially fast, whereas in the second it is polynomially fast, and therefore much slower than in the first case. Notice, however, that at the level of renormalized objects up in the syntax tree, the two situations are completely equivalent, see Fig.\ref{fig11}. This means that the correlation functions for the second sentence (Fig.\ref{fig11}), but at some longer time scales (i.e., at the level of renormalized syntagmas up in the tree), decay exactly in the same way as those in the first sentence (see Fig.\ref{fig10}).

Three remarks are in order. First, in intermediate situations between those described by the two examples above we expect also an intermediate regime between the two limiting cases from Eq.(\ref{de1}) and Eq.(\ref{de2}), but always obeying Eq.(\ref{de0}) asymptotically. Second, notice that the correlation time $\tau$ measures roughly how fast these correlations decay: the shorter $\tau$ is, the faster they decay towards zero. And third, notice that Eqs.(\ref{de0}), (\ref{de1}) and (\ref{de2}) essentially imply that language, at least within this description, has always very short-distance correlations \emph{within the syntax tree}, which does not necessarily imply short-distance correlations in the separation distance within a sentence, { since these two distances are different as shown in Eq.(\ref{de2}). In fact, we will show in Sec.\ref{newsec} that for actual separation distances in the sentence one finds that, on average, language has long-range correlations, i.e., they decay polynomially, in turn matching the experimental observations \cite{experiments}.} Similar conclusions apply as well in the case of having a small number of loops in the network, e.g. in linguistic chains, or in situations such as the German language, where a word correlated with the beginning of the sentence is actually sent to the end.

\subsubsection{Positivity}
By construction, the syntactic TNs presented here are such that all the tensors are non-negative, i.e., they are made entirely of non-negative coefficients. This is because of the stochastic nature of the MERGE tensor, which has been defined in terms of probabilities. These are sometimes called \emph{stochastic tensor networks}. It is well known that such positivity restriction on the coefficients of the tensors is in fact very stringent \cite{positi} and usually implies a very large dimension for the vector spaces of the coarse-grained variables in the TN. We may therefore expect that a TN description of the overall probability distribution in terms of non-negative tensors lowers down this dimension, thus making the representation computationally more efficient. The price to pay, however, is that we loose the interpretation of the MERGE tensor as a tensor of probabilities. Still, a non-positive TN may be computationally more convenient in some situations.   

\subsection{Refinement levels}
The TN structure of MERGE tensors that we just described admits different levels of refinement, when it comes to determining the actual probability of a sentence in a given language model. A practical evaluation of such probabilities, once a parsed corpus (a \emph{Penn TreeBank}) is given, proceeds as follows: 

\vspace{3pt}

(i) First, one does a frequency count of all the words, and computes the probability of being some lexical category ($N$, $V$, etc) conditioned to being a certain word. This probability distribution corresponds, formally, to a MERGE operation at an initial time scale $z_0$ between a set of words and a set of lexical categories, as mentioned briefly in the caption of Fig.\ref{fig3}. In practice, though, it can be accounted for by a $2$-index probability matrix $\widetilde{M}_{\alpha \beta}$, with the first index referring to a particular word, and the second to its lexical category. 

\vspace{3pt}

(ii) Second, one considers every sentence in the corpus and the respective syntax tree, and computes the probabilities corresponding to the coefficients of the MERGE tensors. This is done by counting the frequency of how many times two given lexical elements merge into a given object. Quite importantly, there are (at least) four different levels of refinement of the computed tensors, depending on their position in the $\langle z,t \rangle$ plane and the structure of the syntax tree. In increasing order of refinement, these are: 

\begin{enumerate} 
\item{One single MERGE tensor $M$ for all possible positions in the $\langle z,t \rangle$ plane.}
\item{One MERGE tensor $M^{[z]}$ for each possible renormalization scale, each one for all possible positions in $t$ at the corresponding scale.}
\item{One MERGE tensor $M^{[z,t]}$ for each possible position in the $\langle z,t \rangle$ plane.} 
\item{One MERGE tensor $M^{[T,z,t]}$ for each possible position in the $\langle z,t \rangle$ plane, and for each possible syntax tree $T$.}
\end{enumerate}

The more refined the information included in the computed MERGE tensors, the more accurate is the probability distribution, and therefore the better is the language model. The first of the refinement levels described above corresponds to the probabilistic language models provided by PCFGs \cite{pcfg}. These models are known to work reasonably in some circumstances, although on average not as good as, say, $N$-gram models \cite{ngram}. But this is understandable, because one does not expect a priori the same MERGE tensor at all the possible positions in the $\langle z, t \rangle$ plane. Importantly there are still three more levels of refinement, which should account for better models. The second level drops the assumption of ``ancestor-free" (akin ``scale invariance" in physical jargon), so that the tensors may depend on the scale $z$. The third level drops, additionally, also the assumption of ``place invariance" (akin ``translation invariance" in physical jargon), so that the tensors may also depend on the variable $t$. Finally, the fourth  level of refinement drops the assumption of the MERGE tensors being tree-independent. In principle, the four refinement levels are computable from a TreeBank, implying increasing level of precision for the language model. As for the computational cost of retrieving the MERGE tensors, in the first three levels it should be $O(M \bar{n} )$ both for time and memory, with $\bar{n}$ the average number of words per sentence in a corpus containing $M$ sentences. In the fourth case, however, the time cost is also the same but the memory cost may be larger since, for a large text, we expect to find roughly all possible syntax trees for every sentence length, which for $n$ words is in turn given by the $(n-1)$th Catalan number,  
\beq
C_{(n-1)} = \frac{(2 (n-1))!}{n! (n-1)!} \approx \frac{4^n}{\sqrt{\pi} n^{3/2}} \left( 1 + O\left( \frac{1}{n} \right) \right), 
\eeq
where the approximation is in the limit $n \gg 1$, and therefore scales exponentially. However, typical sentences in human language do not usually imply a dramatically-large number of words (we elaborate more on this in Sec.\ref{sec4}), and therefore the number of different syntax trees to be stored in memory may not be as large in practice as the above number. 

\vspace{3pt} 

Once the MERGE tensors have been computed from the TreeBank, the numerical probability for a sentence of $n$ given words can be obtained in a two-step process: 

\vspace{3pt} 

(i) First, compute the possible syntax trees of the sentence (there may more than one valid tree in ambiguous cases). 

\vspace{3pt} 

(ii) Second, evaluate the probability for each tree following the correlated factorization procedure explained in the previous section, according to the four refinement levels mentioned above. The overall probability is the sum of probabilities for each valid syntax tree. 

\vspace{3pt} 

As the probabilities are computed, it is possible to calculate standard benchmark measures of language models, such as the so-called \emph{perplexity} ${\mathcal P}$, 
\beq
{\mathcal P} = 2^{H(p)} = 2^{- \sum_{\{ w \} } p_{w_1, \cdots, w_n} \log_2 p_{w_1, \cdots, w_n} },
\label{perp}
\eeq
with $H(p)$ the Shannon entropy of the probability distribution. The lower the perplexity, the more peaked is the distribution and thus the better it predicts the sample. So, the better the language model, the lower its  perplexity, at least a priori. In our case we also expect the perplexity to decrease substantially as the refinement of the coefficients of the MERGE tensors increases, according to the four refinement levels mentioned above. Moreover, the perplexity also goes down with the precision of the probabilities in our MERGE tensors. We prove these points in Sec.\ref{qcomp}, using at some steps a novel reformulation of language models in terms of quantum states.

{ 
\section{Long-range correlations in language}
\label{newsec}

How are correlations, on average, for language? It turns out that we can provide an answer for this question using the results that we presented so far. We saw in Eqs.(\ref{de0}, \ref{de1}, \ref{de2}) that correlation functions $C(i,j)$ between two points $i$ and $j$ in a sentence decay exponentially with the syntactic distance $d(i,j)$, i.e., the distane between positions $i$ and $j$ \emph{in the network}. And this distance may be the actual distance in the sentence, as in Fig.\ref{fig10}, or not, as in Fig.\ref{fig11}. 

In practice, it is clear that all languages tend to have more structures like the one in Fig.\ref{fig11} than those like in Fig.\ref{fig10}. The simple reason for this is that MERGE favours the formation of tree-like structures. So, if we were to count the number of possible tree-like sentences in a language (i.e., like those like Fig.\ref{fig11}), they would certainly outnumber the ``linear" structures (i.e., like those in Fig.\ref{fig10}). The conclusion from this is that the average syntactic distance $\overline{d(i,j)}$ over all the sentences of a given language, obeys 
\beq
\overline{d(i,j)} \approx \log_2 \left|j-i \right|, 
\eeq
where the overline means ``average". This must be the case, simply because most of the syntactic structures are trees as in Fig.\ref{fig11}. In turn, this implies that the average correlation function $\overline{C(i,j)}$ behaves like 
\beq
\overline{C(i,j)} \approx \frac{1}{\left| j-i \right|^{1/\overline{\tau}}}, 
\label{corrij}
\eeq 
with $\overline{\tau}$ correlation time that depends only on the average over all sentences of the specific language being analyzed. Such a behaviour is also inherited by other quantities like the \emph{mutual information} $I(i,j)$, since it satisfies the inequality \cite{ignacio}
\beq
I(i,j) \ge a \times C(i,j)^2,  
\eeq
with $a$ some constant prefactor. Thus, on average and for the mutual information one finds  
\beq
\overline{I(i,j)} \ge a \times \frac{1}{\left| j-i \right|^{2/\overline{\tau}}}, 
\label{iij}
\eeq
meaning that it also decays at least polynomially with the separation distance $\left|j - i \right|$ in the sentence, and with a characteristic correlation time $\overline{\tau}/2$. 

The result we just found in Eqs.(\ref{corrij}, \ref{iij}) deserves some discussion. First, notice that we obtained these expressions \emph{exactly}. They therefore imply that languages, on average, have a polynomial decay of correlation functions and mutual information between two points. Importantly, this has been observed in experiments and numerical analysis of specific texts and languages, and is what people often call ``long-range correlations" in language \cite{experiments}.  {Our results thus provide a first intuition on the origin of long-range correlations in language. As we can see here, they are a natural consequence from syntactic trees, since tree-like structures display this type of correlations. In long pieces of texts, it is expected also that the information in sentences organizes also in some sort of hierarchical higher-level structures (sections, chapters, etc) due to semantic/pragmatic constraints. We argue, therefore, that may actually be the reason for long-range correlations observed in long pieces of text including many sentences.} 

Second, notice that some properties of  Eqs.(\ref{corrij}, \ref{iij}) are language-independent, and therefore \emph{universal}, whereas other are language-dependent. In particular, the average polynomial decay is universal. However, the characteristic time scale for this decay, namely exponent $\overline{\tau}$, is language-dependent. The number $\overline{\tau}$ therefore provides \emph{a quantitative way to classify different  texts and different languages}. Those texts and languages with similar syntactic structures will have similar values of $\overline{\tau}$, and is therefore natural to think that they belong somehow to the same family, i.e., probably coming from the same ancestor (e.g., Spanish, Italian and French all coming from Latin). And this is important, because such an analysis may be a way to provide hints about the genealogical (or ``genetic") root of  languages for which we do not know their origin  \cite{langisolate}, such as Basque, Sumerian, and others.   
}

\section{Language model quantum states} 
\label{qcomp}

Let us now define the following quantum state: 
\beq
\ket{\Psi(T_n)} = \frac{1}{Z(T_n)^{\frac{1}{2}}} \sum_{w_1, ..., w_n} \left(p_{w_1, \cdots, w_n}\right)^{\frac{1}{2}} \ket{w_1, \ldots, w_n},
\label{langstate}
\eeq
with $p_{w_1, \cdots, w_n}$ the probability of a sentence with words $w_1, \cdots, w_n$ and syntax tree $T_n$, and $\{ \ket{w_1, \ldots, w_n} \}$ an orthonormal (tensor product) basis of some Hilbert space for $n$ parties, each party corresponding to the position of a word in the sentence. The dividing normalization factor $Z(T_n)$ is actually the partition function of the probability distribution, i.e., 
\beq
\braket{\Psi(T_n)}{\Psi(T_n)} = \frac{1}{Z(T_n)} \sum_{w_1, ..., w_n} p_{w_1, \cdots, w_n} = 1. 
\eeq
We call the state in Eq.(\ref{langstate}) a \emph{language model quantum state}. 

\begin{figure}
	\centering
	\includegraphics[width=0.4\linewidth]{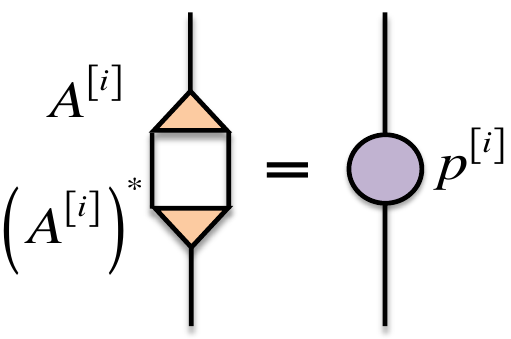}
	\caption{(Color online) TN diagram for Eq.(\ref{proba}). The matrix on the right hand side is diagonal, and with entries $p^{[i]}_\gamma \delta_{\gamma \gamma'}$.}
	\label{notiso}
\end{figure}

Because of the correlated factorization of syntactic TNs explained in previous sections, one can see easily that these  language model quantum states admit a TN representation of their coefficients, i.e., they are really \emph{TN states} in the strict quantum-mechanical sense. The TN structure of the coefficient $\left(p_{w_1, \cdots, w_n}\right)^{\frac{1}{2}}$ is simply given by the same one as for the probability distribution $p_{w_1, \cdots, w_n}$ (the syntactic TN), but replacing every coefficient of a MERGE tensor by its square root. More specifically, it is the same TN but with $3$-index tensors $A^{[i]}$ of coefficients
\beq
A^{[i]}_{\alpha \beta \gamma} \equiv \left(M^{[i]}_{\alpha \beta \gamma} \right)^{\frac{1}{2}}, 
\eeq
again with $i$ simply label for the different tensors. This simple prescription is a direct consequence of tensors being diagonal in the syntactic TN. Notice also that these tensors obey the condition 
\beq
\sum_{\alpha, \beta} A^{[i]}_{\alpha \beta \gamma} \left( A^{[i]}_{\alpha \beta \gamma'}\right)^* = \left( \sum_{\alpha, \beta} M^{[i]}_{\alpha \beta \gamma} \right) \delta_{\gamma \gamma'} = p^{[i]}_\gamma \delta_{\gamma \gamma'}, 
\label{proba}
\eeq
with $p^{[i]}_\gamma$ the probability of merging at position $i$ any two given lexical objects into $\gamma$, and $\delta_{\gamma \gamma'}$ the Kronecker delta, see Fig.\ref{notiso}. 

\subsection{Properties} 

The language TN quantum state that we just defined is interesting for a number of reasons. These  are described in what follows. 

\subsubsection{Truly random sampling}
First, notice that if this quantum state becomes (somehow) experimentally available in an actual quantum system, then it can be used to do \emph{truly random sampling} of the probability distribution of sentences with that particular syntax tree. For comparison, all classical samplings are based on pseudo-random number generators, which are known to induce errors in the long run for, e.g., Monte Carlo methods. The state could also be useful, for instance, to find the most-likely sentences in a language model, and things alike. Notice that, here, we understand by ``perfect sampling" the fact that the sampling does not introduce spurious correlations caused by a pseudo-random number generator. Still, near-term quantum devices are known to be noisy, which could also affect this property if implemented in  practice. 

\subsubsection{Language model quantum circuit} 

\begin{figure}
	\centering
	\includegraphics[width=0.9\linewidth]{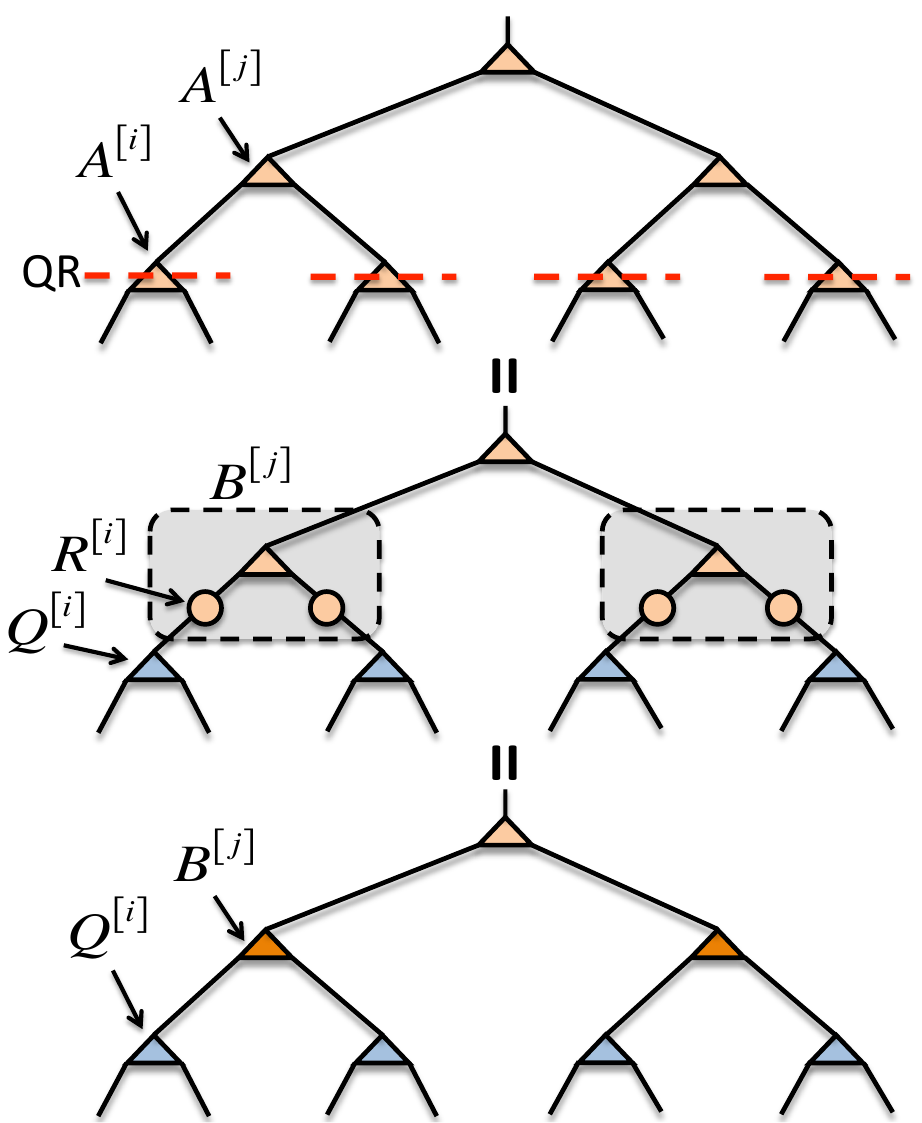}
	\caption{(Color online) Iterative procedure to get the quantum circuit producing a language model quantum state for a given syntax tree (see text). The red dashed lines in the upper diagram correspond to $QR$ decompositions. The process is iterated at every scale, until reaching the top.}
	\label{qcirc}
\end{figure}

\begin{figure}
	\centering
	\includegraphics[width=0.9\linewidth]{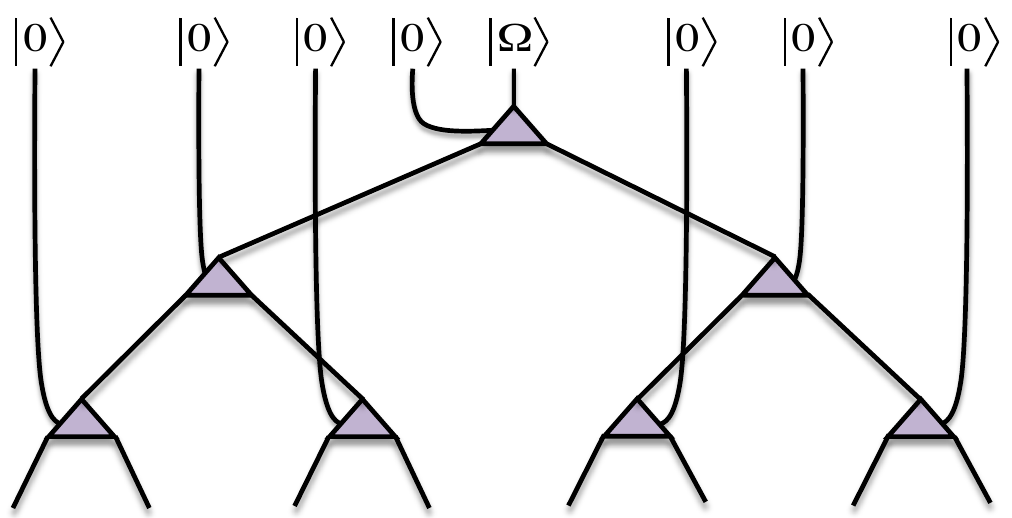}
	\caption{(Color online) Quantum circuit of $2$-body gates producing a language model quantum state for a given syntax tree. Ancillary degrees of freedom are fixed to the quantum state $\ket{0}$. The state $\ket{\Omega}$ at the top may be produced from $\ket{0}$ by some extra $1$-body gate, and its squared norm codifies the overall probability of the tree.}
	\label{qcirc2}
\end{figure}

Second, the state can, in fact, be created by a \emph{quantum circuit} with as many two-body gates as $A$-tensors. The procedure is sketched in Fig.\ref{qcirc}: starting from the shortest renormalization scale $z_1$, one reshapes the indices of the $A$-tensors as a matrix and performs a $QR$ decomposition \cite{qr}, as shown in the figure. Since the $A$-tensors are real and positive, the matrix $Q$ is orthogonal, i.e., $Q^T Q = \mathbb{I}$. Reshaping back $Q$ as a $3$-index tensor provides an isometric tensor, which we keep at the particular sites of the network at that renormalization scale. Matrices $R$, however, are contracted with the $A$-tensors at the next renormalization scale $z_2$, see Fig.\ref{qcirc}. The resulting tensors, call them $B$, are then also $QR$-decomposed, where the $Q$s define again isometries, which we keep in the network, and the $R$s are contracted with the $A$-tensors at the next renormalization scale. By iterating this process up to the top level, one gets a TN of \emph{isommetric} $3$-index tensors $Q^{[i]}$, and a quantum state $\ket{\Omega}$ at the very top carrying non-local information about the probability of the whole sentence. In particular, since tensors $Q^{[i]}$ are isommetries, one has that 
\beq
\braket{\Psi(T_n)}{\Psi(T_n)} = \frac{1}{Z(T_n)} \braket{\Omega}{\Omega} = 1, 
\eeq
(where the last equality follows from the normalization of the state), and therefore
\beq
\braket{\Omega}{\Omega} = Z(T_n) = \sum_{w_1, ..., w_n} p_{w_1, \cdots, w_n}, 
\eeq
which means that the norm of the quantum state $\ket{\Omega}$ is the overall probability of having an $n$-word sentence (whichever) with syntax tree $T_n$ in the language model. This global information just moved up to the top level of the TN and, importantly, we constructed it locally at every renormalization scale by a sequence of $QR$ decompositions,  therefore very efficiently (notice that we \emph{never} needed to compute each one of the terms $p_{w_1, \cdots, w_n}$ individually!) \footnote{This is in fact a very efficient procedure to compute the overall probability of a given tree in a language model.}. Connecting to the usual developments in quantum-mechanical TN states, this is an example of an isometric TTN state \cite{ttn}. Finally, in order to promote this structure to a quantum circuit, we simply notice that an isometric tensor can be understood as a two-body unitary gate, where one of the indices is fixed to some ancillary state $\ket{0}$ \cite{er}, see Fig.\ref{qcirc2}. The resulting diagram is nothing but the picture of the quantum circuit producing the desired quantum state. The conclusion is that if the MERGE tensors are given, then one could in principle produce these quantum states efficiently in a quantum computer or a quantum simulator. Last but not least: the description above has been for TNs without loops, but it can be generalized to other situations. In case of having a small number of loops in the network (e.g. in CHAINS), there is also a similar procedure as the one indicated here by playing with several tensor decompositions ($QR$, Singular Value Decomposition, etc), always sending the non-unitary parts upwards in the syntactic network. { We envisage that this may trigger applications of near-term quantum processors for computational tasks related to language.} 

\section{Perplexity from entanglement} 
\label{secperplexity}

An interesting application of the language model quantum states defined in the previous section  concerns the perplexity ${\mathcal P}$ of a language model, which was defined in Eq.(\ref{perp}). For a given sentence, it turns out that we can give lower bounds on the perplexity of a given subset of words, using tools from quantum information theory, as we show next.  

\begin{figure}
	\centering
	\includegraphics[width=0.8\linewidth]{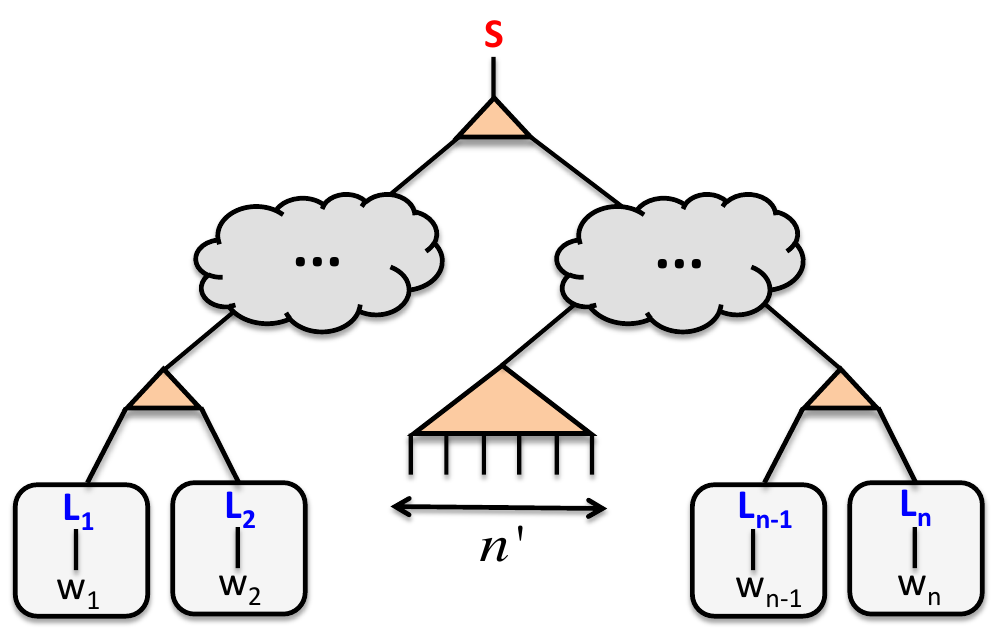}
	\caption{(Color online) Subset of $n'$ contiguous words in an arbitrary sentence, as described by the language quantum state in Eq.(\ref{langstate}). The clouds indicate some arbitrary piece of an arbitrary syntactic tree.}
	\label{rho1}
\end{figure}

\begin{figure}
	\centering
	\includegraphics[width=1\linewidth]{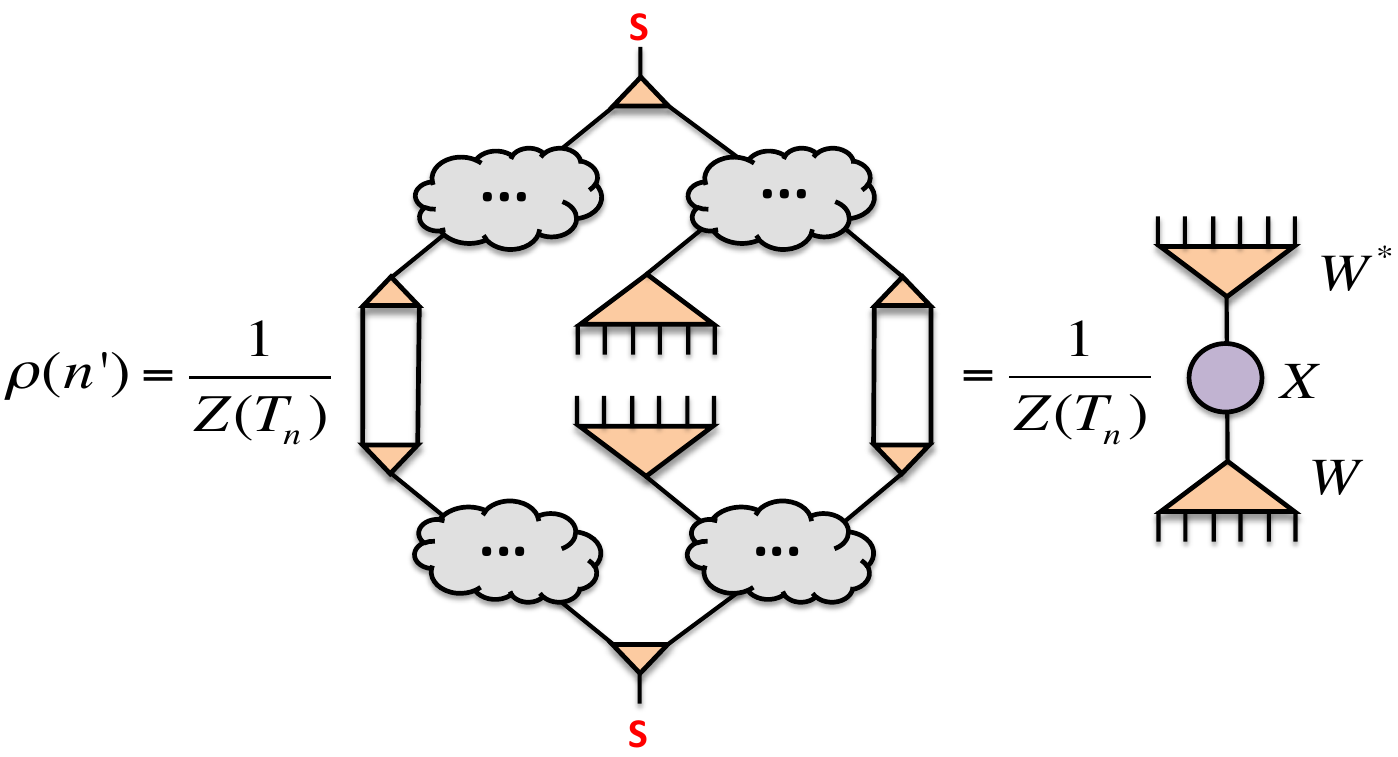}
	\caption{(Color online) Reduced density matrix of a block of contiguous $n'$ words in the language state of Eq.(\ref{langstate}).}
	\label{rho2}
\end{figure}

\begin{figure}
	\centering
	\includegraphics[width=1\linewidth]{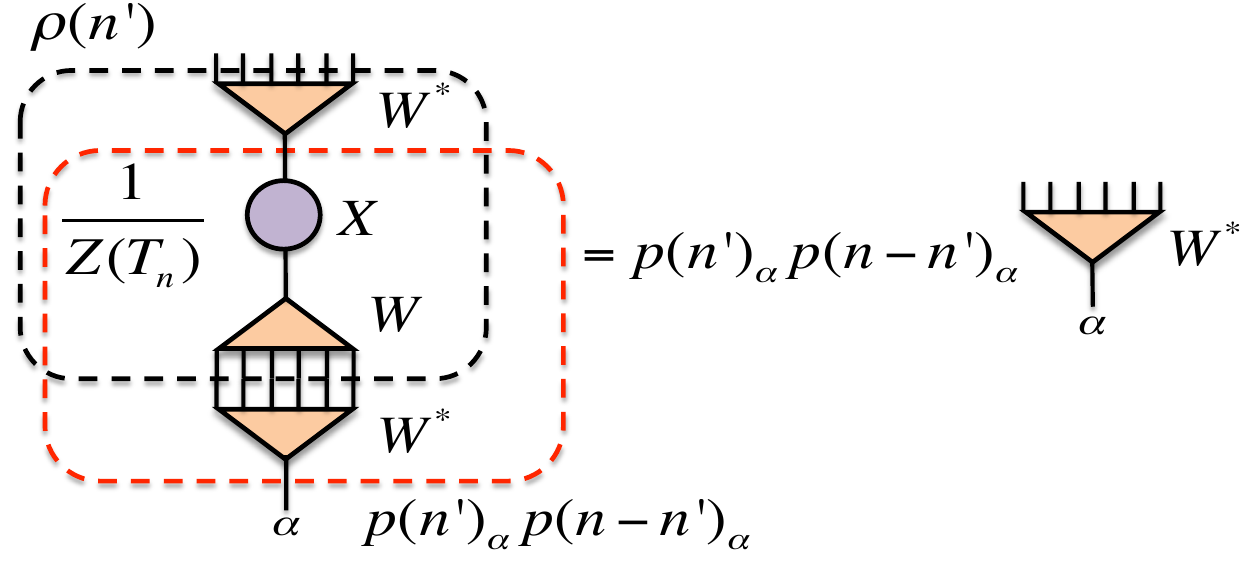}
	\caption{(Color online) TN diagram for the eigenvalue equation of the reduced density matrix $\rho(n')$.}
	\label{rho3}
\end{figure}

Let us start by considering a sentence with $n$ words, and a subset of $n' < n$ contiguous words within the sentence. These are a \emph{block} of $n'$ words. The question we want to answer now is: how much is the entanglement of this block of $n'$ words in a given quantum state $\ket{\Psi(T_n)}$ for a syntax tree $T_n$? Following the usual procedure  for bipartite entanglement, we get first the reduced density matrix of the block, 
\beq
\rho(n') = {\rm tr}_{n-n'}  \ket{\Psi(T_n)} \bra{\Psi(T_n)} ,  
\eeq
with ${\rm tr}_{n-n'} ( \cdot )$ the partial trace over the rest of the system (the \emph{environment}). As shown in the diagrams of Fig.\ref{rho1}, this can be achieved by ``cutting" out the relevant sub-tree linking the $n'$ words from the rest of the sentence. After the appropriate contractions, this reduced density matrix can always be written as 
\beq
\rho(n') = \frac{1}{Z(T_n)} W X W^{\dagger} 
\eeq
with $W$ some rectangular matrix amounting for the contraction of the sub-tree for the block, and $X$ a square matrix whose rank is the number of lexical categories $N_l$ in our grammar, being this also the rank of $\rho(n')$, see Fig.\ref{rho2}. It is easy to see, moreover, that in fact matrix $X$ is diagonal, 
\beq
X_{\alpha \alpha'} \propto p(n-n')_\alpha \delta_{\alpha \alpha'}, 
\eeq
with $p(n-n')_\alpha$ the overall probability of the string of $n-n'$ words merging into lexical category $\alpha$, no matter the words in the string. One can also see that the (unnormalized) eigenvectors of $\rho(n')$ are given by 
\beq
\left(v_\alpha \right)_\omega = \left( W^{\dagger} \right)_{w \alpha}, 
\eeq
with $\left(v_\alpha \right)_\omega$ the $\omega$-coefficient of the $\alpha$th eigenvector, and eigenvalues $\lambda_\alpha$ given by 
\beq
\lambda_\alpha = p(n')_\alpha  ~ p(n-n')_\alpha, 
\label{eig}
\eeq
with  $p(n-n')_\alpha$ as described above, and similarly for $p(n')_\alpha$ but for the set of $n'$ words, see Fig.\ref{rho3}. Using Eq.(\ref{eig}), one can get the \emph{entanglement entropy} $S(\rho(n'))$ and the \emph{single-copy entanglement} $E_1(\rho(n'))$ of the block of $n'$ words \cite{entanglement}, which are given respectively by 
\beqa
S(\rho(n')) &=& - \sum_\alpha \lambda_\alpha \log_2 \lambda_\alpha \nonumber \\
E_1(\rho(n')) &=& - \log_2 \left( \max_\alpha \lambda_\alpha \right).  
\label{scopy}
\eeqa
The above entanglement measures obey the chain of inequalities
\beq
E_1(\rho(n')) \leq S(\rho(n')) \leq \log_2 N_l, 
\label{ine}
\eeq
which implies that the entanglement of the block can never be too large, since the number of lexical categories $N_l$ in a typical grammar for human language is usually quite small.

Next, we notice that the probability distribution $p_\omega$ for the $n'$ words in the block is actually given by the diagonal elements of $\rho(n')$ in the basis of Eq.(\ref{langstate}) restricted to the block, i.e., 
\beq
p_\omega  = \rho(n')_{\omega \omega}.  
\eeq
One can check from the derivations above that this probability distribution and the one of the eigenvalues $\lambda_{\alpha}$ obey the majorization relation \cite{majorization} 
\beq
\vec{p} \prec \vec{\lambda},  
\eeq
which implies 
\beq
H(p_w) \ge S(\rho(n')), 
\eeq
i.e., the Shannon entropy of the reduced probability distribution for the block of $n'$ words is larger than the entanglement entropy of the block. This relation, combined with Eq.(\ref{ine}), implies directly that 
\beq
{\mathcal P} = 2^{H(p_w)} \ge 2^{S(\rho(n'))} \ge 2^{E_1(\rho(n'))},  
\eeq
with ${\mathcal P}$ the perplexity of the distribution of the $n'$ words as defined in Eq.(\ref{perp}).  Combining this with Eq.(\ref{eig}) and Eq.(\ref{scopy}), in the end we arrive to the result 
\beq
{\mathcal P} \ge \min_\alpha \left( \frac{1}{ p(n')_\alpha  ~ p(n-n')_\alpha} \right),  
\label{bound}
\eeq
which is our main lower-bound for the perplexity of the probability distribution of the $n'$ words. 

Some remarks are in order. First, notice that Eq.(\ref{bound}) is a fully classical result, even though we used the machinery of quantum information theory to find it. Second, the inequality is giving us a fundamental lower bound on how well our language model can predict sentences, just because of its statistical nature. Third, we can roughly estimate the scaling of this lower bound: if $p_{\rm max}$ is the maximum merging probability over all MERGE tensors in the network, it is easy to see that 
\beq
{\mathcal P} \gtrapprox \left( \frac{1}{p_{\rm max}} \right)^{n-1} 
\eeq
which implies, also roughly,  that the perplexity gets worse (increases) exponentially fast with the number of words $n$ in the sentence, but also that it improves (decreases) exponentially fast if the MERGE probabilities of the language model get more refined and accurate. This inequality shows clearly the route required in order to improve the performance of syntax-based probabilistic language models: short sentences, and accurate probabilities.  

\section{Arbitrary grammars and language models} 
\label{depen}

\begin{figure}
	\centering
	\includegraphics[width=0.9\linewidth]{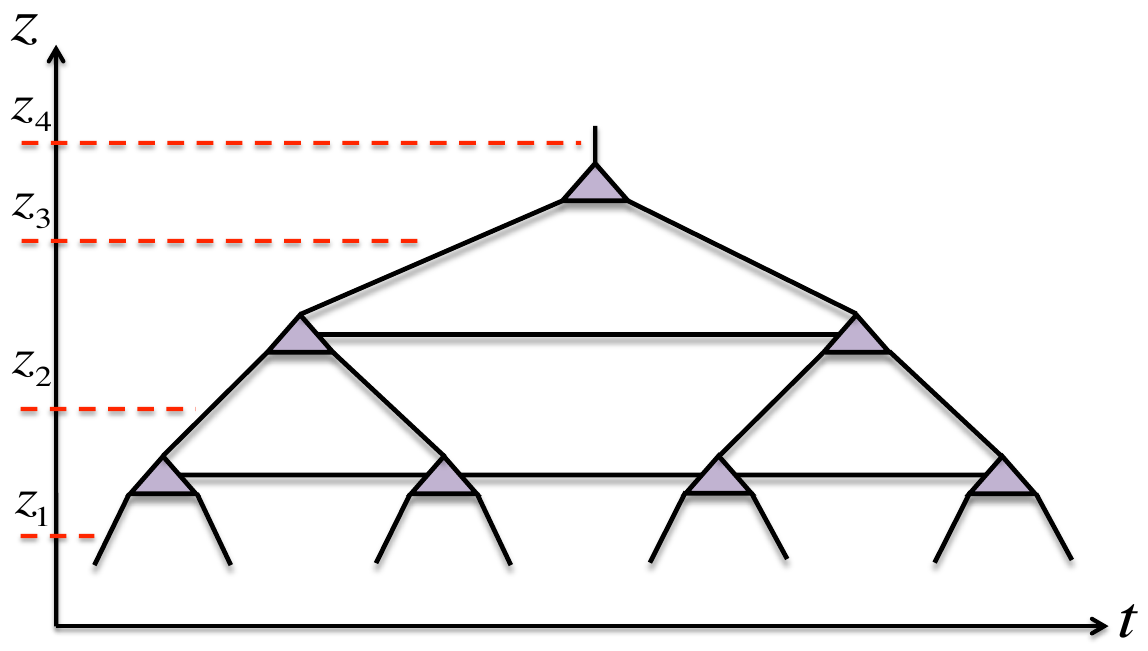}
	\caption{(Color online) Possible MERA-like TN for some possible dependency grammar. Probability distributions (tensors) are correlated at every renormalization scale. The structure is no longer a tree if all possible dependencies are taken into account at every scale, as shown in the diagram.}
	\label{mera}
\end{figure}

We would like to say a couple of words about other types of grammars, not necessarily context-free, as well as other language models. Importantly,  the tensor network picture of language is not necessarily restricted to the cases that we presented above, and in fact can be used to describe the correlation structure of, essentially, any type of grammar and/or language model. For instance, the trees of \emph{dependency grammars} \cite{dependen}, though not based on the MERGE operation, also admit a TN representation of their correlations when put as a probabilistic language model. We could even add long-range dependencies between the probability distributions in constituency grammars, as was shown for the case of chains in Fig.\ref{fig8}, but which can in fact be generalized over the whole $\langle z,t \rangle$ plane, obtaining what is known in physics as a MERA-like tensor network \cite{er}, see Fig.\ref{mera}. As a matter of fact, it would be possible to model with TNs \emph{any} grammatical correlation structure, even if not directly linked to human language. An example would be a syntactic structure based on an hypothetical MERGE operation with multiple outputs for a given input. Such structures would not have the property of ``correlated factorization" discussed above, but most of the key properties that we mentioned would still hold, including those related to computational  efficiency and short-range syntactic correlations. 

\begin{figure}
	\centering
	\includegraphics[width=0.6\linewidth]{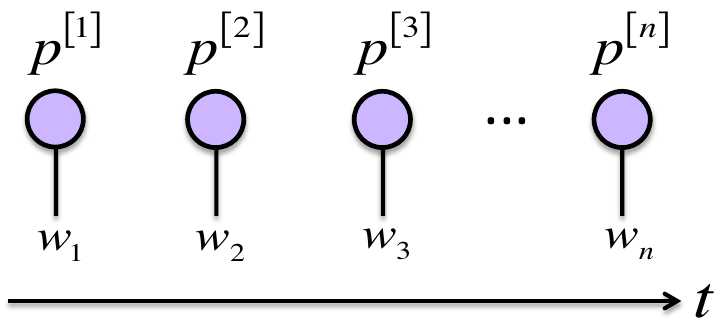}
	\caption{(Color online) TN for a $1$-gram language model. Only the time axis is relevant, and there is no correlation between the words $w_1, ... , w_n$. In physics, this is the analogue of the so-called mean-field theory approximation.}
	\label{fig9}
\end{figure}

From a practical perspective, the so-called \emph{$N$-gram models} \cite{ngram}, where the probability of observing a word is assumed to depend only on the history of the preceding $N-1$ words, also admit a similar description. For instance the case of $1$-grams corresponds to the product probability distribution
\beq
p_{w_1, \ldots, w_n} = p^{[1]}_{w_1} \cdots ~ p^{[n]}_{w_n}, 
\eeq
which can be represented by the TN diagram of Fig.\ref{fig9}. Such a $1$-gram TN does not include \emph{any} correlation between the words. For comparison, similar separable TNs are also the ones used in the so-called \emph{mean-field} approximation to strongly correlated systems, where correlations between different sites are discarded \cite{mf}, and which is known to fail whenever correlations are important. For the case of more complicated $N$-grams, one can actually define an appropriate language model quantum state, i.e., 
\beq
\ket{\Psi(N-{\rm gram})} = \frac{1}{Z^{\frac{1}{2}}} \sum_{\alpha \in {\rm N-gram}} \left(p_{\alpha}\right)^{\frac{1}{2}} \ket{\alpha},
\eeq 
with $\alpha$ an index running over all possible $N$-grams, $p_{\alpha}$ their probabilities, $\ket{\alpha}$ a set of orthonormal states, one for every $N$-gram (which is rather easy to construct), and $Z$ the partition function of the distribution. Once such a state is available, one can do similar things as for the TN language models discussed previously, such as truly random sampling, and so forth.

\section{Implications}
\label{sec4} 

Our ``renormalization picture" of syntax and the results presented above demand for a necessary and detailed discussion about its implications, which extend into different ambits. In what follows we elaborate on some of them, taking a somehow more phylosophical perspective than in the previous sections, though well-grounded in our rigorous observations so far. 

{ 
\subsection{Good and bad models for language processing}}
The first practical implication, as we have already hinted in the previous sections, is that ``good"  language models (of any kind) should be compatible with the coarse-graining picture that we presented. From a generic perspective, one should expect a language model to reproduce the way humans seem to organize correlations in sentences, and from our perspective, this is given by the organization of coarse-grained information at different time scales. Concerning the field of artificial intelligence, we thus believe that a good starting point to obtain better language-processing algorithms, is to include also this organization of linguistic information according to time scales. This is in fact partially achieved already by the so-called ``syntactic language models" \cite{synt}. The same applies to theoretical models of language in theoretical linguistics \footnote{A recent attempt in this direction, also related to quantum physics and linear algebra, is the Matrix Syntax model in Ref.\cite{MS}.}. Notice, importantly, that in this work we never hypothesized about what is the fundamental theory of grammar behind the known properties of MERGE. Questions such as ``why a noun and an adjective merge into a noun phrase?", or ``why is the output of MERGE uniquely determined by its input?", are beyond the scope of this work. In other words: we observed how correlations in human language get organized, explained this organization using the tools of physics, and exploited the consequences. And that is everything we did.  We never discussed where these correlations could come from, or why they are as they are. In any case, and this is the point that we wish to make here, models attempting to explain this, either computational or theoretical, should encompass the picture presented here to be legitimate, since our observations are general.  

{ 
We are now in position to answer the following question: why are some neural networks good at processing language, whereas some others are bad? The reason for this is now clear: those neural networks that are good, such as deep convolutional networks, are nothing but Tree Tensor Networks (TTNs)  \cite{neural}, and therefore codify the correct renornalization structure. As we explained before,  TTNs can encompass the long-range correlations observed in language (i.e., the polynomial decay of correlation functions and of mutual information for two words in a sentence, see Sec.\ref{newsec}).  Additionally, those neural networks that are bad at language processing, such as recurrent neural networks and hidden Markov models, also have a TN structure \cite{neural} but it corresponds to an MPS, i.e., a structure such as the one in Fig.\ref{fig10}. As we explained also before, MPS have exponentially-decaying correlation functions and thus cannot account for the average long-range correlations of language. The message is then clear: some neural networks work well because they have the correct renormalization structure (e.g., deep convolutional networks), whereas other work badly because they do not have such a structure (e.g., recurrent neural networks and hidden Markov models). In fact, the connection between neural networks and renormalization had already been pointed out, see for instance Ref.\cite{roinick}. Notice that, as a byproduct, our observation implies that one could have discovered the structure of \emph{biological} neural networks not by looking at how the actual neurons are organized in our brains, but rather by analyzing the correlation structure of the language output of these brains. From such an analysis we could have already concluded, as the only logical option, that something like neurons must be somehow interconnected in a way compatible with the renormalization of information. As we know, this is in fact the case, and it is nothing but what tries to be mimicked by artificial neural networks. The point is, that one could have hinted this biological structure without having to open any brain!} 
 
{ 
\subsection{Universal and non-universal properties}
}

Given the renormalization structure and the properties of TN language models, one can predict universal features, i.e., properties that should be the same, no matter the language, and which only depend on the correlation structure of syntax \footnote{In words of N. Chomsky, ``there is only one human language" (private communication).}. {  In this paper we found already some of such universal properties, as discussed in Sec.\ref{newsec}: the polynomial decay of correlation functions and mutual information is universal. However, the characteristic time-scale $\overline{\tau}$ for this decay is language-dependent and should depend on external factors, such as cultural heritage. }

We wish to remark that universal properties of language had already been observed by analyzing linguistic information with the tools of \emph{complex networks} \cite{cn}. This is the field of physics and mathematics that analyzes complex systems and their structure from the network perspective (examples are ubiquitous: the internet, the power grid of a country, financial networks, the synaptic network in the brain...). In this setting, the so-called \emph{linguistic networks} allow for a study of the properties of syntax from a pure network-theory perspective. In order to avoid confusion, we stress that our approach here is radically different, since we start from a very different physical perspective: renormalization, and how this orchestrates correlations. This led to a TN picture of language models which is different but complements the one obtained using complex-network theory.  

\bigskip

\subsection{Optimality of language} 

Several of the properties from the previous section seem to be related to the conjectured ``perfection and economy" of human language in the Minimalist Program, as well as to the conjectured efficient processing of linguistic information in the brain \cite{MP, effbrain}.  Let us take for concreteness the language models that we analyzed before. The fact that the TN structures are mostly loop-free automatically implies that the retrieval of information can be done efficiently in all computational resources (a problem in the complexity class P). Such computational efficiency strongly depends on the quasi-loop-free renormalization structure of syntax trees, and is therefore generically valid, i.e., not just for the case of language models. In fact, loop-free structures are well-known to be the \emph{cheapest non-trivial} class of correlation structures in terms of the manipulation of their information \cite{ttn}. The surprising fact is that human language is \emph{even more efficient than this}, because of the properties of MERGE. In particular, we saw that the uniqueness of the output of MERGE once the input is specified, implied diagonal tensors and thus a correlated factorization in the TN, which leads to a dramatic efficiency in the calculation of probabilities for TN language models. It looks, therefore, that the human language chose the cheapest possible option able to keep non-trivial correlations between information units. Our brain could have evolved to use a MERGE where the output is non-unique for a given input and still maintain a big part of the computational efficiency in the manipulation of information, but this just didn't happen. This observation makes precise the common-lore statement that language is, indeed, the cheapest non-trivial computational system. This may be one of the reasons why our brains choose to work with such correlation structures, instead of a different one. And we manage to externalize it through a physiological interface pretty well: we communicate (on average and most of the time) via sequential sounds in time produced with one mouth, instead of producing correlated sounds with, say, each one of our fingers, which would amount to 20 mutually correlated outputs, and thus a syntax full of correlation loops in turn implying computational inefficiency in the processing of its information. 

\subsection{Non-Markovian memory environment}

A coarse-graining is a process that finds effective degrees of freedom to describe an emergent object, and inherently involves an \emph{information loss} when moving from one scale to the next. It is well known in physics that renormalization is, usually, irreversible (the so-called ``irreversibility of RG flows") \cite{iRG}. In language, however, it is clear that even if syntax manipulates coarse-grained objects at some long-time scale, we still \emph{know} about the information content of the short-time scales. This is, our brain seems to organize the information according to different time scales, but does not seem to fully \emph{erase} the information when going from one scale to the next, at least for some period of time. For instance, when we say a sequence of the type $[_{NP} ~ [_A ~ X] ~ [_N ~ Y ]]$ (an adjective $X$ followed by a noun $Y$), we remember for a while what it actually refers to:  ``happy cat",  ``hot meal", ``interesting paper", and so on. This seems to indicate that the ``discarded" information seems not to be immediately erased, but just put apart for a while in some memory degree of freedom. To put it in physical jargon, one would say that the ``memory environment" is \emph{non-Markovian}, in the sense that there seems to be access for some period of time to the discarded information, shall this be needed. Understanding how and why this happens is indeed a relevant but different question to the one that we addressed in this paper.  

\subsection{Context-free grammars in other ambits}
An interesting observation is that probabilistic context-free grammars (PCFG), though originally developed in linguistics, have proven recently very powerful in the probabilistic modelling of RNA and protein structures. In particular, PCFGs offer a way of determining the secondary structure of RNA, with a comparable accuracy to that obtained by energy minimization methods \cite{rnasec}. Concerning proteins, the situation is more complex but several achievements have already been reported using PCFG methods \cite{prosec}. Many of the things that we mentioned previously in this work for the case of  language, therefore, apply as well to the study of RNA and protein sequences. Even if being a very different scenario, the relevant correlation structures that appear in these biological problems happen to be similar to the ones that we described  in this work, and therefore the same derivations could be applied to study those. The same is also true for the correlation structures present in \emph{programming languages}, such as C++, Java, and so on. From a theoretical perspective, programming languages actually apply the rules of some grammar, i.e., rules by which words in a computer code are interpreted into meaningful machine instructions.  

Intriguingly, one can also make a turnaround in the derivation that we presented here, and consider some TN structures as the natural correlation output of grammars.  To be more precise, one could argue that TTNs and MPS can, in general, always be regarded as the output of some set of ``generalized" context-free grammar rules where one allows for several possible outputs of a MERGE operation for a given input, being the outputs associated to complex ``weigths". As such, this then implies that ground states of gapped $1d$ local quantum many-body Hamiltonians, which are known to have an MPS structure \cite{tn},  are (roughly speaking) nothing but generalized grammatical structures. Whether this simple observation has consequences in the (analytical and numerical) study of quantum many-body systems remains as a provocative open question. 

\subsection{On typical human abilities}
Intriguingly, similar structures to the ones presented here for the case of language and grammar have also been found in different but somehow related scenarios. For instance, as we said before,  the correlation structure of neural network algorithms (which mimic in part the behavior of neurons in the brain) is, in fact, that of a Tree Tensor Network \cite{neural}. Renormalization-like algorithms are also common in the study of image compression, such as those based on wavelets \cite{wavelets}, and even on Matrix Product States \cite{ji}, where information of a picture gets organized according to different $2d$ length scales. Matrix Product States have also been used in the context of machine learning \cite{miles}. Moreover, it has been argued that the harmonic structure of tonal music may be, in fact, also a result of the MERGE syntactic operation  \cite{music}. As a matter of fact, it is believed that the faculty of language appeared in evolution almost simultaneously to the faculties of mathematics and music, with some people arguing in favour of the three faculties being actually three different manifestations of the same basic ability, which became available to our ancestors due to some genetic mutation throughout evolution \cite{three}. A very subtle, somehow missed point, but key in this regard, is that the mathematical faculty looks itself \emph{also} as a coarse-graining of (mathematical) information. This is in fact a consequence of MERGE being the successor function in mathematics \cite{succ}. In order to make this point more explicit, let us directly cite a rather popular paragraph (at least in the linguistics' community) in one of the recent works of N. Chomsky \cite{onphases}: 

\begin{quotation}
\emph{``Suppose that a language has the simplest possible lexicon: just one lexical item, call it ``one". Application of MERGE to the lexical item yields \{one\}, call it ``two". Application of MERGE to \{one\} yields \{one, \{one\}\}, call it ``three". And so on. In efect, MERGE applied in this manner yields the successor function. It is straightforward to define addition in terms of MERGE(X,Y), and in familiar ways, the rest of arithmetic. The emergence of the arithmetical capacity has been puzzling ever since Alfred Russell Wallace, the co-founder of modern evolutionary theory, observed that the ``gigantic development of the mathematical capacity is wholly unexplained by the theory of natural selection, and must be due to some altogether distinct cause", if only because it remained unused. It may, then, have been a side product of some other evolved capacity (not Wallace's conclusion), and it has often been speculated that it may be abstracted from the faculty of language by reducing the latter to its bare minimum. Reduction to a single-membered lexicon is a simple way to yield this consequence."}
\end{quotation}

Moreover, and at an experimental level, neuroscientists have recently discovered what could be the signature of the MERGE operation in neural activity, by analizing the neural activation of epileptic patients performing several language tasks \cite{pnas}. 

Given all this, we take the liberty to take off and hypothesize, somehow phylosophically and because everything seems to point in this direction, that the human abilities of language, mathematics, and probably others, may actually be different manifestations of a fundamental single ability of the human brain, namely, \emph{the ability to organize and process information according to different physical scales}. { To put it simple: one could say that \emph{the human brain is, among other things, a biological information-renormalization machine.}} When it comes to human language, this allows the brain to build a \emph{language system of discrete infinity}, i.e., a discrete and recursive system able to produce infinitely-many outputs. 

\bigskip 

\section{Conclusions and perspectives}
\label{sec5}
The observations and results in this paper are highly interdisciplinary. Let us briefly summarize here the main points. We have argued that the linguistic MERGE operation entails renormalization in physics: the information content in, e.g., sequences of words (short time scale) gets renormalized by MERGEs up to sentences (long time scale). We have made this observation {exact and formal} for language models, and have found that probabilities of meaningful sentences are naturally given by quasi-loop-free TNs, which in turn organize correlations according to different renormalization time scales. Such language models are naturally related to probabilistic context-free grammars, though not restricted only to them.  We have discussed some of the properties of these TN language models: locally-built syntactic correlations at every scale, very high efficiency of information processing because of correlated factorization of the TN, syntactic correlations, and practical refinement levels. { We also proved that long-range correlations in language follow naturally from our approach.} Moreover, we proposed how to promote probabilistic language models to probability distributions of quantum states, argued that such quantum states may be useful when it comes to sampling the distribution, showed how they can be built efficiently in a quantum computer, and used their entanglement properties to provide a classical lower bound on the statistical perplexity of finding a set of words in a sentence. We discussed also how this useful formalism may be generalized to other types of grammars, and discussed a number of implications of our observations in several ambits. These concern the legitimacy of language models { and neural networks for language processing,} universality and optimality of language, some required properties of the memory environment, the potential application of our formalism for RNA and protein sequencing as well as programming languages and quantum many-body systems via context-free grammars, and the overall picture of several human faculties all somehow boiling down to MERGE. In the end, we have taken the liberty to hypothesize that the human brain seems to have a natural fundamental ability to organize information according to different physical scales, from which other faculties may materialize. 

Our work opens the possibility to use all the mathematical and physical knowledge about TN states, both classical and quantum, in the theoretical and computational study of language and grammar. This includes a wide variety of applications not just in linguistics, but also in RNA and protein sequencing \cite{rnasec, prosec} and the design of computer languages, just to name some well-known examples. In particular, the different ways to quantify correlations and the information content in the network, as well as associated numerical algorithms \cite{tn}, should find useful applications in these scenarios. Moreover, the efficient descriptions of probability distributions of relevant grammars by means of quantum states, opens the exciting possibility to use possible quantum computers and quantum simulators to deal with problems in all these ambits. A prominent example is AI, where our results show that quantum information tools can be used to validate, simulate, assess, and improve state-of-the-art language models, as well as that quantum computers can be used to implement perfect random sampling of language, which is simply impossible with classical technology. This is particularly relevant given the recent big advances in the development of experimental quantum processors. Additionally, it would also be interesting to study the connection between the observed structure of probability distributions for language, and those obtained from the renormalization method for reaction-diffusion models in physics \cite{reaction}, which seem to share some similarities.  

By digging deeper into linguistic concepts it is indeed possible to take our {analogies} further. We do this in Appendix \ref{appa}. All in all,  our conjecture that MERGE in linguistics is connected to RG in physics turns our to be extremely fruitful, since many of the key linguistic ideas from the last century fit perfectly with known physical concepts linked to renormalization. We have also seen that, as a consequence, many concepts in computational linguistics also match perfectly with well-known physical conceptions. The main {analogies}  discussed in this paper, including those in the appendix, are summarized in Table \ref{tabsumm}. 

\begin{table}[h]
\centering
	\begin{tabular}{|c|c|} 
	\hline 
	~~~Linguistics~~~ & ~~~~~~~Physics~~~~~~ \\
         \hline 
         \hline
         MERGE & Coarse-graining \\ 
	Relabelling & Rescaling \\
	Derivation & RG flow \\ 
        Phase & RG scale \\
        Phase impenetrability  & RG irreversibility\\
        Optimality and efficiency & Loop-free structures \\  \hline \hline 
        Prob. language model & 1d tensor network \\
        $N$-gram models & Mean-field theory \\
        ~Prob. context-free grammar~ & 3-index tensor \\
        & $\&$ MPS/TTN \\
        Dependency grammar & $(k>3)$-index tensor \\
        & $\&$ 1d MERA \\
        $\sqrt{{\rm Prob.~ language ~ model}}$ & Quanum circuit \\
        Perplexity & ~Quantum entanglement \\ \hline 
        \end{tabular} 
	\caption{Main {analogies} and connections between linguistics and physics proposed in this paper. The upper part corresponds to concepts usually discussed in theoretical linguistics, and the lower part to concepts in computational linguistics. 1d means that the ``physical" degrees of freedom span along one dimension, which in the case of language is time. The ``square-root" symbol in the lower-left panel is a way of saying that the corresponding quantum circuit produces probability amplitudes that are the square root of the actual probabilities given by the language model.}
\label{tabsumm}
\end{table}

Only good things can happen by studying language from the perspective of physics. Our work here is one more addition to the long-standing program of using physics tools to understand the linguistic capacity. Other works have focused on descriptions based on quantum field theory, complex network theory, and quantum mechanics \cite{more}, and the results in this paper add a fresh new perspective to this overall framework. The fields of physics and linguistics have been traditionally very far away from each other. But indeed, linguistics focuses on the study of the laws of language, and physics on the study of the laws of Nature. For a linguist, the human language is the universe, and it has deep connections with how our brain processes and manipulates information, as well as other situations whose correlations are orchestrated by grammar-like rules.  From the perspective of physics, it feels just natural to think that classical and quantum information theories should be somehow useful for this purpose. Being able to formalize mathematically some of the most relevant aspects of language and grammar in terms of physical ideas is already an  important achievement. We strongly believe that the cross-fertilization of physics and linguitics will become increasingly relevant in the future. Phylosophical questions, such as those encountered sometimes in linguistics, usually lead to deep, profound scientific problems, and our work here is no exception to this rule.  

\acknowledgments 
We acknowledge the Universitat Aut\`onoma de Barcelona, the Max Planck Institute of Quantum Optics, and the University of Mainz, where the ideas in this paper materialized from interdisciplinary discussions  over several years. Discussions with Ondiz Aizpuru, Gustavo Ariel Schwartz, Gemma de las Cuevas, Laura Garc\'ia-\'Alvarez, Geza Giezdke, Jos\'e I. Latorre, Enrique Solano, Miles Stoudenmire and Juan Uriagereka are acknowledged. A special acknowledgement to Sergi Quintana, for explaining us the basics of Chomsky's generative grammar 28 years ago: { \emph{qui sembra, recull.}}

\appendix 

\section{More {analogies} by digging deeper}
\label{appa}

Our paper is written having in mind a reader with background on physics and mathematics. However, the topic itself is strongly interdisciplinary. Because of this, in this appendix we would like to add some extra information useful for the reader with knowledge about theoretical linguistics. In particular, we would like to define a few concepts more precisely in linguistic jargon. Thanks to this, we will see that  by digging deeper into the linguistic jargon, more {analogies} with physics will show up, in turn strengthening our thesis that MERGE in linguistics and RG in physics are deeply linked to each other.  

To begin with, the term \emph{Universal Grammar} (UG) is nothing but a label for the striking difference in cognitive capacity between ``us and them", i.e., humans versus the rest of animal species. UG is thus the research topic of generative grammar in its attempt to understand what it is and how it evolved in our species. Finding a satisfying answer to the latter question may be impossible with the tools we have right now, but any theory of UG seeking to address the former must meet a criterion of evolvability: any properties, mechanisms, etc. attributed to UG should have emerged in what appears to have been a unique and relatively sudden event on the evolutionary timescale \cite{evolut}. This line of thought presupposes that UG (the genetic encoding of the the human linguistic capacity) manifests \emph{bona fide} traits of perfect design, in the sense that contains operations and mechanisms that follow from conceptual necessities, efficiency principles or interface demands. In this respect, linguistic expressions (sentences, phrases, words, etc.) are built up by adhering to these principles, therefore in an optimal fashion. While these notions are intuitively clear,  their precise formulations remain vague and controversial. 

One of the most important mathematical achievements of generative grammar is the so-called ``Chomsky Hierarchy" \cite{Chomsky1956}, a classification of formal grammars according to their complexity. As Chomsky showed sixty years ago, human languages manifests both context-free and context-sensitive properties, needed to construct PHRASES and CHAINS respectively, shown in the Sentences \ref{s1}(a,b): 
\beqa
\label{s1}
&a.& {\rm John ~ killed ~ John.} \\
&b.& {\rm John ~ was ~ killed ~ <John> } \nonumber
\eeqa
In Sen.\ref{s1}(a) (a PHRASE) we have two tokens of the lexical item ``John" that participate in phrasal dependencies to yield a compositional interpretation whereby the first John is the agent of a killing event, and the second John is the patient of such event. What we have in Sen.\ref{s1}(b) (a CHAIN) is more complex. This time, we don't have two tokens of ``John", but two occurrences of the same lexical item -- as if they were one and the same object in two positions at the same time, where the notation ${\rm <John>}$ means that the word itself is not pronounced at that position, but it is also interpreted there from the logical point of view. This is what is called CHAIN in linguistics. In languages of the English type, the first (leftmost) occurence is spelled-out, whereas the second (rightmost) is necessary to keep a syntax-semantics homomorphism (that is, to capture the desideratum that a specific interpretation is tied to a specific position). Notice that the same type of object (a CHAIN) is necessary in Sen.\ref{s2}, where ``John" is pronounced to the left of \emph{seem}, although it is interpreted as the patient of \emph{killed}.
\beq
{\rm John ~ seems ~ to ~ have ~ been ~ klilled ~ <John> }
\label{s2}
\eeq

In order to account for these properties, generative grammar has resorted to phrase structure rules (PSR) and transformations. The most articulated version of PSR is known as $X$-bar Theory, which resorted to different devices that have been subject to a revision within minimalism. In particular, Chomsky \cite{merge} argued that the basic properties of PSR could be understood by means of a computational operation, dubbed MERGE \cite{merge}, which captures two empirical properties of human language that are non-negotiable: \emph{discrete infinity} and \emph{displacement}. To be able to account for those properties, one must assume an operation that constructs hierarchically structured expressions with displacement. And that is what MERGE does. 
MERGE applies to two objects $X$ and $Y$ (be these words or bigger units), yielding a new one, $K$, which is the set containing $X$ and $Y$, i.e., $\{ X, Y \}$. If $X$, $Y$ are distinct (taken directly from the \emph{lexicon} or independently assembled), $K$ is constructed by what is called EXTERNAL MERGE (EM); if $Y$ is part of $X$ (if $Y$ is contained in $X$), then we have what is called  INTERNAL MERGE (IM). The latter scenario is that of Sentences \ref{s1}(b) and \ref{s2} above, where MERGE turns ``John" into a discontinuous object (a CHAIN). For completeness, if the operation is at the beginning of a derivation (e.g., with bare lexical items from a lexicon), it is called FIRST MERGE, and if it operates with partially-derived items (phrases), it is called ELSEWHERE MERGE. 

Chomsky \cite{merge} takes MERGE to be strictly binary, as it is what is minimally necessary to create hierarchical structure. Generation by MERGE thus entails a restrictive class of recursively defined, binary-branching and discrete-hierarchical structures.  

It is also worth mentioning that in $X$-bar Theory, the label identifies the properties of the entire phrase, at the cost of this being a theory-internal symbol that departs from inclusiveness demands. An alternative to this is a label-free representation (see Fig.\ref{fig1}), where  endocentricity (the assumption that all phrases must be headed) is not preserved. This entails that syntactic objects can be exocentric, as seems to be necessary for objects formed by the combination of two phrases, $\{XP,YP \}$. Syntactic objects are ``endocentric" if they contain an element that can be determined by Minimal Search -- typically, a head. Given this logic, $\{X,YP \}$ is endocentric and $\{XP,YP \}$ exocentric. Consequently, such a system freely generates objects of different kinds, without stipulating their endocentric nature.
	
Moreover, MERGE is subject to efficiency and economy conditions. One such condition is inclusiveness, which precludes the introduction of extraneous objects, like the ones that $X$-bar Theory deployed: traces, bar-levels, projections, etc. Inclusiveness also bars introduction of features that are not present in lexical items.

To further clarify MERGE, we stress that the combination of two objects, $X$ and $Y$, yields a new one, $K$, which is the set $\{ X, Y \}$. Once we have $\{X,Y \}$, we may want to merge $K$ and some object $W$, which can be either internal to $K$ or external to it (see above). In any event, the merger of $W$ cannot change or tamper with $\{X,Y \}$, which behaves as a unit. More precisely, subsequent applications of MERGE must yield Eq.\ref{s3}(a), not \ref{s3}(b):
\beqa
\label{s3} 
&a.& {\rm MERGE} (K,W) = \{ \{ X, Y \},W \} \\
&b.& {\rm MERGE} (K,W) = \{ \{ X, W \},Y \} \nonumber
\eeqa

The driving force of this work is the fact that MERGE and renormalization seem to play a similar role on various respects. As noted above, MERGE takes two objects, $X$ and $Y$, to yield a new one, $K$, thus removing $X$ and $Y$ from the computational workspace (WS). In the simplest scenario, MERGE maps ${\rm WS} = [X,Y]$ onto ${\rm WS'} = [\{X,Y \}]$, reducing the complexity of WS. Notice that MERGE never extends the WS, at least in terms of cardinality; thus ${\rm WS} = [\{X,Y \}]$ and ${\rm WS'} = [\{W,\{X,Y\}\}]$ are equally bigger, since they only contain one set. A new element can be added to WS (or ${\rm WS'}$) in only one way: by taking two items $W$, $Z$ from the lexicon and introducing $\{W,Z \}$ into WS as a new element, yielding ${\rm WS''} = [\{W,Z\},\{X,Y\}]$. Of course, cardinality can be reduced if we apply EM (EXTERNAL MERGE) and neither of the elements are taken from the lexicon, as if we map ${\rm WS''} = [\{W,Z\},\{X,Y\}]$ onto ${\rm WS'''} = [\{\{W,Z\},\{X,Y\}\}]$. This idea is indeed very similar to that of a coarse-graining in physics, in the sense made precise throughout the paper. 

Additionally, the possibility that computational load is reduced by MERGE is perhaps somewhat new, as this typically follows from a principle in linguistics that is called STRICT CYCLICITY. The notion of cycle (and thus cyclicity) goes back to the fifties, where work in phonology \cite{cyc} showed that the application of stress-assigning rules apply from innermost to outermost units of a word, putting aside linear order information. More generally, an object is build under cyclic principles if it is COMPOSITIONAL, which means that its interpretation is fixed by the elements it contains and the way in which they are combined. Consider this with Sentences \ref{sc}, where the interpretation is crucially different (Brutus is an agent in (a), and a patient in (b)), although both examples contain the same three words:
\beqa
\label{sc}
&a.& {\rm Brutus ~ stabbed ~ Caesar.} \\
&b. & {\rm Caesar ~ stabbed ~ Brutus.} \nonumber
\eeqa

The concept of STRICT CYCLICITY is a stronger version of cyclicity. The key intuition behind it is that for certain linguistic object constructed in a derivation (say, a $VP$), further computation should not modify it. Let us see this with the example in Eq.\ref{sd}, where the verb ``leave" is merged with the $NP$ ``the room" to yield the complex $VP$ ``leave the room", which we can call $K$ for ease of reference.
\beq
\label{sd} 
{\rm MERGE}({\rm leave,\{the,room \}}) = {\rm \{leave,\{the,room\}\}}
\eeq

What is of interest here is that the interpretation of $K$ (that is, of ``leave the room") is determined at that stage of the derivation (at that ``cycle"), and cannot be changed at subsequent stages (``cycles"). Therefore, if we add ``Mary" to obtain ``Mary leaves the room" (call it $K'$), as in Eq.\ref{st}, the interpretation of $K$ will be the same in Eq.\ref{sd} and in Eq.\ref{st}. 
\beq
\label{st} 
{\rm MERGE} ({\rm Mary},K) = {\rm \{Mary,\{leaves,\{the,room\}\}\} }
\eeq

In a nutshell, the interpretation of complex objects is constructed stepwise, in a step-by-step fashion, and whatever has been done at a stage $s$ cannot be undone at statge $s+1$ (Eqs.\ref{sd} and \ref{st} above). This, in turn, is quite analogue to the idea of irreversibility of RG flows in physics, which matches perfectly with our interpretation of MERGE as a coarse-graining of information.  

Such stages at a derivation, where a ``computation" is done and cannot be altered afterwards, correspond with the so-called linguistic PHASES, and the device responsible for ensuring that the interior of a PHASE is no longer accessible is the PHASE IMPENETRABILITY CONDITION (PIC for short). What has been called ``phase" roughly corresponds with the notion of ``cycle" described above. Using the physical interpretation that we introduce in this paper, one would say that a PHASE in linguistics is the analogous of an RG scale in physics. 

To be more precise, a phase is defined in linguistics as a domain $D$ where uninterpretable features (number and person features of verbs) are valued. When a phase is closed off, the complement domain $\Omega$ (which can itself be complex, in the sense of having some inner structure) of the phase head $P$ cannot be modified from the outside; this means, for instance, that the case of an $NP$ within $\Omega$ (e.g., ``the book" in the $VP$ ``read the book") cannot be changed once the phase headed by $P$ is complete \cite{phaseapp}. Among other things, this entails that ``the book", which is the Direct Object of ``read" Sentence \ref{sp} (it receives accusative case from ``read"), cannot also be the Direct Object of the matrix verb ``believe":
\beq
\label{sp} 
{\rm I ~ believe ~ that ~ John ~ read ~ the ~ book.}
\eeq

That ``the book" is the Direct Object of ``read" and not of ``believe" is shown in Sentences \ref{sw},  where we see that this $NP$ can be passiviced in the embedded clause, but not in the matrix clause ($*$ signals ungrammaticality):

\beqa
\label{sw} 
&a.& {\rm I ~ believe ~ that ~ the ~ book ~ was ~ read.} \\
&b.& {\rm *The ~ book ~ was ~ believed ~ that ~ John ~ read.} \nonumber
\eeqa

This ``shielding" effect that makes the $VP$ impenetrable is captured by the phase impenetrability condition mentioned above. Physically, this is the irreversibility of the RG flow when moving from one RG scale to the next.  There are various approaches to Phase Theory \cite{Gallego2012}, but all of them share the key intuition that PHASES are domains where complexity is reduced by somehow allowing the system to ``forget" about an amount of structure that has been created and which will be no longer accessible. This process of ``forgetting" is, in fact, analogous to the process of ``discarding irrelevant degrees of freedom" in an RG-step in physics. 

Moreover, the ``rescaling" step in RG has not been discussed in this paper, but also appears naturally when particularizing to specific models of language. For instance, in the Matrix Syntax model \cite{MS} this rescaling appears naturally in order to recover the correct linguistic labels after a MERGE operation (see Ref.\cite{MS} and the discussions therein for more information). We believe that this is a general feature: the ``rescaling" in physics is nothing but the ``mathematical relabelling" that one needs in order to recover the correct labels ($NP$, $VP$, etc) after a MERGE operation when dealing in practice with models of language.

{\bf Funding:} This project was not funded by any grant. 
 
{\bf Conflict of interest:} The authors declare that they have no conflict of interest.

\end{document}